\documentclass[preprint,times,review]{elsarticle}

\usepackage{graphicx}
\usepackage{amsmath}
\usepackage{amssymb}
\usepackage{booktabs}

\usepackage{multirow}
\usepackage{bm}
\usepackage{color}

\usepackage{algorithm} 
\usepackage{algorithmic} 

\usepackage{subfigure}
\usepackage[normalem]{ulem}
\useunder{\uline}{\ul}{}

\newenvironment{packed_itemize}{
	\begin{itemize}
		\setlength{\itemsep}{1pt}
		\setlength{\parskip}{0pt}
		\setlength{\parsep}{0pt}
	}{\end{itemize}}

\usepackage{xspace}
\makeatletter
\DeclareRobustCommand\onedot{\futurelet\@let@token\@onedot}
\def\@onedot{\ifx\@let@token.\else.\null\fi\xspace}

\def\eg{\emph{e.g}\onedot} 
\def\ie{\emph{i.e}\onedot}

\def\etal{\emph{et al}\onedot}
\makeatother

\usepackage{amsmath}

\usepackage{booktabs,subcaption,amsfonts,dcolumn}
\newcolumntype{d}[1]{D..{#1}}

\usepackage[pagebackref,breaklinks,colorlinks]{hyperref}

\usepackage[capitalize]{cleveref}
\crefname{section}{Sec.}{Secs.}
\Crefname{section}{Section}{Sections}
\Crefname{table}{Table}{Tables}
\crefname{table}{Tab.}{Tabs.}

\journal{Pattern Recognition}

\begin{document}
	\captionsetup[figure]{name={Fig.}}
	\begin{frontmatter}
            \title{Improving Deep Representation Learning via Auxiliary Learnable Target Coding}
    
            \author[pcl,scut]{Kangjun Liu}
            \ead{liukj@pcl.ac.cn}
            \author[pcl]{Ke Chen\corref{cor1}}%
            \ead{chenk02@pcl.ac.cn} 
            \author[scut]{Kui Jia}%
            \ead{kuijia@scut.edu.cn}
            \author[pcl]{Yaowei Wang}
            \ead{wangyw@pcl.ac.cn}
    	\cortext[cor1]{Corresponding authors.}
     
            \address[pcl]{Pengcheng Laboratory, Shenzhen 518000, China }      
            \address[scut]{South China University of Technology, Guangzhou 510641, China}
    		
    	\begin{abstract}
                Deep representation learning is a subfield of machine learning that focuses on learning meaningful and useful representations of data through deep neural networks. However, existing methods for semantic classification typically employ pre-defined target codes such as the one-hot and the Hadamard codes, which can either fail or be less flexible to model inter-class correlation. In light of this, this paper introduces a novel learnable target coding as an auxiliary regularization of deep representation learning, which can not only incorporate latent dependency across classes but also impose geometric properties of target codes into representation space. Specifically, a margin-based triplet loss and a correlation consistency loss on the proposed target codes are designed to encourage more discriminative representations owing to enlarging between-class margins in representation space and favoring equal semantic correlation of learnable target codes respectively. Experimental results on several popular visual classification and retrieval benchmarks can demonstrate the effectiveness of our method on improving representation learning, especially for imbalanced data.
    	\end{abstract}
    	\begin{keyword}
                Image classification, Representation learning, Target Coding.  
    	\end{keyword}
        \end{frontmatter}

\section{Introduction}\label{sec:introduction}
In the past decade, deep representations have been verified their superiority to hand-crafted representations owing to encoding optimal features from data in an end-to-end learning manner. 
In the context of supervised visual classification, the one-hot code is commonly used under the ideal assumption of a balanced data distribution of available training samples, which can be significantly invalidated in practice.
As a result, learning discriminative representations from imbalanced visual data remains active since the rise of deep learning.

To address such a challenge, most of the existing algorithms are concerned with improving feature discrimination via data augmentation as implicit regularization \cite{lin2024good} and explicit feature regularization \cite{Ioffe2015BatchNA}, owing to their capability of preventing from over-fitting.
Alternatively, a number of recent methods \cite{liu2022convolutional, yang2021learning} were proposed to learn better feature representations by constructing latent correlations between different categories.
By utilizing such target relation, minority classes can leverage cross-class dependencies to share information with majority classes, thus mitigating the effects of data imbalance when training deep models that rely solely on category labels.

In addition, previous studies such as \cite{Yang2015DeepRL, yuan2020central} have shown that using multiple positive elements in target codes for supervised classification can outperform the one-hot code. 
{Specifically, an information-rich target code \cite{Yang2015DeepRL} can significantly enhance the feature representation during deep model training.}
This motivates us to investigate the idea of using one type of target codes as an auxiliary regularization term to complement another style of target codes with latent dependency and geometric properties in the target space.
One example is to use Hadamard codes directly \cite{Hedayat1978HadamardMA} as auxiliary supervision signals for classification in addition to the conventional one-hot codes.
Although combining complementary target codes can benefit from enriching the diversity of target coding, using pre-defined target codes is still less flexible in modeling complex semantic relations between target classes.

\begin{figure}[t]
    \centering
    \includegraphics[width=0.75\linewidth]{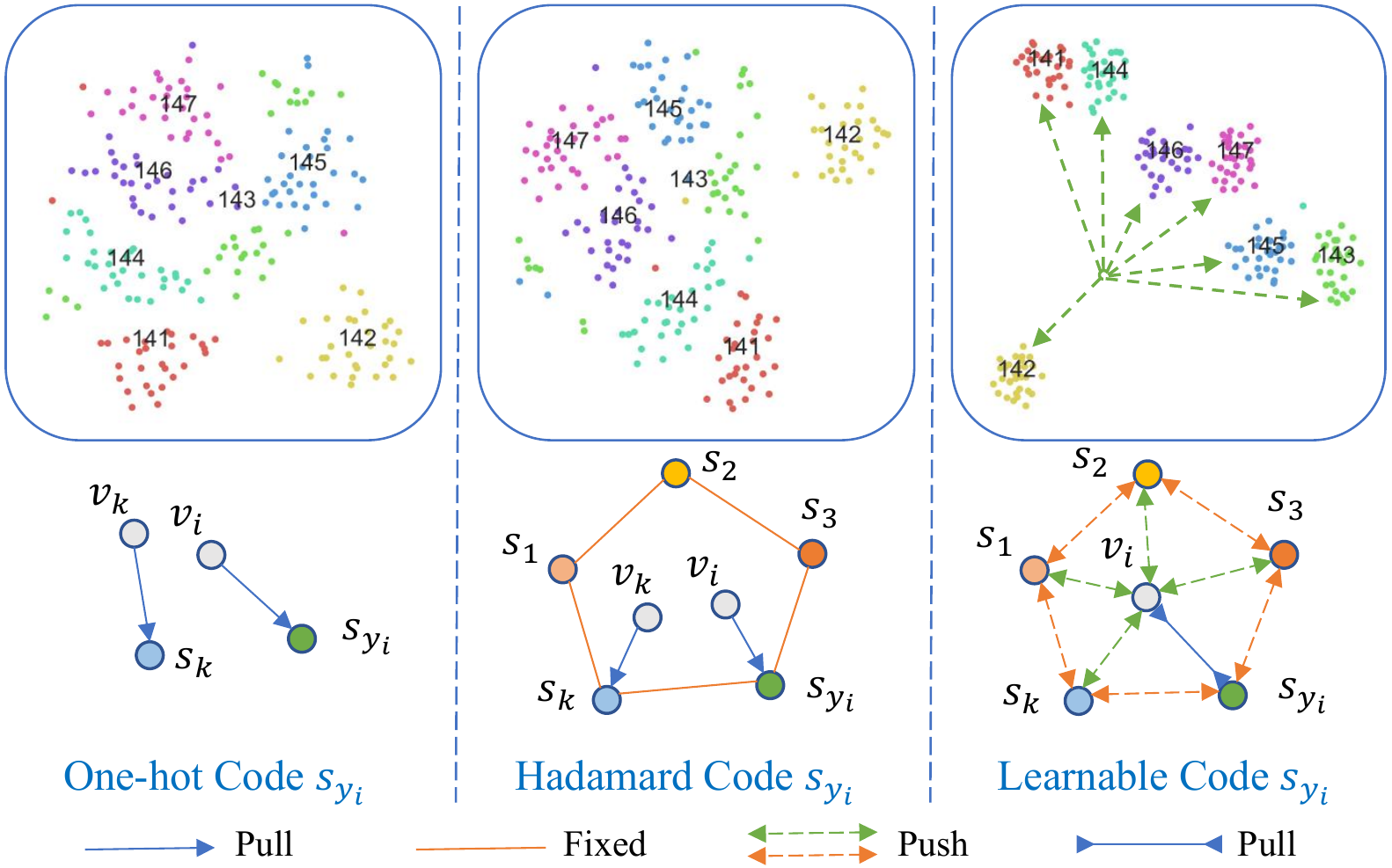}
    \vspace{-1mm}
    \caption{Comparison with the one-hot code, the Hadamard code, and the proposed Learnable code. The projected representations on the top are extracted from the same last block of ResNet backbone of all three methods by the t-SNE \cite{Maaten2008VisualizingDU}, where the classes id 141-147 represent seven different kinds of Terns from the CUB benchmark \cite{wah2011caltech}. It is observed that the model trained with our proposed LTC can gain better representations owing to discovering inter-class relation (see \cref{subsec:LTC}) and also imposing geometric properties of target codes (see \cref{subsec:semantic-correlation}), {which thus demonstrates the effectiveness of our proposed LTC method in making sample features more compact within the same category while increasing the difference between different classes. {Note that, the symbol "$v$" and "$s$" represent the semantic and target code vectors, respectively.}}}
    \label{fig:figure1}
    \vspace{-1mm}
\end{figure}

Therefore, a more flexible target coding method with learnable parameters is proposed in this paper to expand the target representation space.
It is worth noting that the goal of using learnable target codes is similar to that of using Hadamard codes as auxiliary supervision signals. 
However, Hadamard codes cannot be adjusted during model training and require a large number of redundant code dimensions, which can be costly. 
In contrast, our proposed method allows for more flexibility by using learnable parameters and constructing class-wise target codes that reveal the geometric structure in the target space.
Note that, our proposed learnable target codes also bear resemblance to the method presented in \cite{yang2021learning}, which introduced trainable target codes solely for hashing retrieval purposes.
The key distinction between both methods lies in that our proposed learnable target codes are utilized as an auxiliary regularization technique for deep representation learning, coupled with our development of multiple effective loss functions to enhance the learning of these target codes.

Concretely, we design a margin-based triplet loss and a correlation consistency loss to enhance the discriminative power of our deep representations, \ie to increase the distance between representations of samples from different categories and reduce the distance between those from the same categories.
Moreover, our learnable target codes enable the capture and encoding of target correlation across different categories into the code vectors, which, in turn, guide deep representation learning to reflect the structural dependencies present in the feature space.
To better illustrate our motivation, we present a comparison of three types of target codes in \cref{fig:figure1}, which shows that Hadamard codes can slightly compact the representation, while our proposed learnable target codes are more effective in achieving both compactness and relation encoding.

In this paper, we make the main contributions as follows. 
\vspace{-2mm}
\begin{packed_itemize}
\item First, we propose a generic auxiliary feature regularization of deep representation learning by using other complementary target codes, which can be readily applied to existing representation learning algorithms.  
\item {Second, we propose a learnable target coding (LTC) scheme that incorporates a global margin-based triplet loss and a semantic correlation consistency loss. This scheme aims to capture target correlations flexibly and enforce geometric properties, ultimately enhancing deep representations.}
\item Finally, we conduct extensive experiments on three widely-used fine-grained classification datasets, {as well as two fine-grained image retrieval benchmarks and three popular imbalanced image classification datasets to verify the effectiveness of our proposed method. Our method consistently achieves competitive or superior performance to the state-of-the-art on all the benchmarks and datasets.}
\end{packed_itemize} 
\vspace{-2mm}

Source codes are made publicly available at \href{https://github.com/AkonLau/LTC}{https://github.com/AkonLau/LTC}.
\vspace{-2mm}
\section{Related work}\label{sec:relatedWorks}
\vspace{-2mm}This section provides an overview of three related research topics: deep representation learning in fine-grained image recognition, imbalanced data learning, and target coding methods for general representation learning. 
\vspace{-2mm}

\subsection{Deep Representation Learning}
The problem of deep representation learning in fine-grained image recognition \cite{mao2023multi, chen2024fet} as an important sub-problem of visual recognition has been actively studied for decades. 
Existing works can be roughly organized into three categories -- fusion-based methods, attention-based methods, and regularization-based methods. 

\noindent\textbf{Fusion-based --}
Fusion-based methods usually adopt high-order feature fusion operations to improve the discrimination of sample representations.
With Lin \etal \cite{lin2015bilinear} first proposed a bilinear pooling model for learning discriminative features, its follow-uppers \cite{Gao2016CompactBP} propose a compact bilinear pooling method for reducing the dimension of high-order representation, and Yu \etal \cite{yu2022efficient} proposed a more efficient compact bilinear pooling method with a two-level Kronecker product.
Apart from the pooling operation, Qi \etal \cite{qi2019exploiting} proposed to concatenate features corresponding to different spatial relations for enhancing the fusion of features.

\vspace{0.1cm}
\noindent\textbf{Attention-based --}
Attention-based methods are commonly created with the purpose of detecting distinctive regions and acquiring discriminative features by mimicking the human attention mechanism.
Zheng \etal \cite{zheng2017learning} put forward a multi-attention CNN (MA-CNN) framework for both part localization and discriminative feature learning in a weakly supervised and end-to-end manner.
Similarly, He \etal \cite{he2018fast} introduced a weakly supervised discriminative localization (WSDL) method that utilizes a multi-level attention extraction network and multiple localization networks to enable rapid fine-grained feature learning.

\vspace{0.1cm}
\noindent\textbf{Regularization-based --}
Regularization-based methods usually regularize the model training via adopting self-supervised relations or structures. 
Zhou \etal \cite{zhou2020look} proposed an object-extent learning module and a spatial context learning module for modeling the structure of objects and enhancing discriminative feature learning in a self-supervised way.
As an improvement, Wang \etal \cite{p2pnet2022} proposed a self-supervised pose alignment method for learning more discriminative part features. 
Recently, Liu \etal \cite{liu2022convolutional} proposed a dynamic target relation graph, as a self-generated target coding, to promote the model to capture the correlation among different categories. 

Regarding fine-grained representation learning, our proposed method belongs to the category of regularization-based techniques with the same objective as \cite{liu2022convolutional}.
The key distinction lies in the fact that our proposed regularization, based on target coding, can capture category correlations through learnable parameters, rather than solely relying on feature similarity distance as in \cite{liu2022convolutional}.

\subsection{Imbalanced Data Learning}\vspace{-1.5mm}
In recent years, imbalance learning \cite{liu2023noise, zhang2024few} as a widely-encountered problem in real-world data has been explored by numerous studies. 
Lin \etal \cite{Lin2017FocalLF} proposed a dynamically scaled cross-entropy loss (Focal Loss) for dealing with the class imbalance in object detection, where the foreground and background classes are extremely imbalanced.
To better mitigate the class imbalance, Cui \etal \cite{Cui2019ClassBalancedLB} proposed a re-weighting strategy using the effective number of samples for re-balancing the loss function.
In addition, Cao \etal \cite{Cao2019LearningID} proposed to regularize the minority classes with a label-distribution-aware margin (LDAM) loss function and a deferred re-weighting (DRW) training schedule. 
Except for the above loss re-weighting methods, Kang \etal \cite{Kang2020Decoupling} decoupled the model training into the representation learning process and classification process, and found out that strong re-balancing ability can be gained by only adjusting the classifier. 

Compared to the aforementioned imbalanced learning methods, our proposed auxiliary target coding constraint can be regarded as a feature regularization technique that employs class-wise correlation assistance to improve imbalanced data learning.
\vspace{-2mm}

\subsection{Target Coding}\vspace{-1.5mm}
Except for typical one-hot or multi-hot {target codes} in supervised model training, there exist amounts of non-typical target coding methods as supervision.
{To be clear here, ``target code" refers to a specific type of coding scheme, while ``target coding" is used to describe the process of encoding target information in a particular way.}
Earlier works related to target coding methods can be tracked back to pre-defined Target Coding \cite{Yang2015DeepRL}, {weakly-supervised DCE \cite{li2018deep}, semi-supervised S2LFS \cite{li2021semi},} and self-supervised DTRG \cite{liu2022convolutional} for representation learning, CSQ \cite{yuan2020central}, DeepBit \cite{lin2018unsupervised}, AdaLabelHash \cite{yang2021learning} and HashNet \cite{cao2017hashnet} for image retrieval, and Label Smoothing (LS) \cite{Szegedy2016RethinkingTI} and Online Label Smoothing (OLS) \cite{zhang2020delving} for soft target coding, which can be organized into two categories: hand-crafted and learnable.

\noindent\textbf{Hand-crafted --}
Hand-crafted target coding methods \cite{Yang2015DeepRL, yuan2020central} are originally inspired by the Hadamard matrix \cite{Hedayat1978HadamardMA}, where each row generated with binary values always has the same distance among each other.
Target Coding \cite{Yang2015DeepRL} firstly utilized the generated Hadamard codes for generating the deep representation learning just with a Euclidean loss, while CSQ \cite{yuan2020central} adopted the Binary Cross Entropy with a quantization loss for constraining the hash codes learning. 
Label Smoothing (LS) \cite{Szegedy2016RethinkingTI} is a pre-defined soft label technique that has been widely used in deep supervised learning to enhance representation performance, while Online Label Smoothing (OLS) \cite{zhang2020delving} as an improvement is generated through the iterative of training samples' prediction for capturing more statistical information.

\noindent\textbf{Learnable --}
Learnable target code algorithms usually involve defining the target codes using trainable parameters or learning the target codes using deep neural networks. 
For example, HashNet \cite{cao2017hashnet} proposed a novel deep architecture to learn exactly binary hash codes from imbalanced similarity data by continuation method with convergence guarantees.
DeepBit \cite{lin2018unsupervised} proposed an unsupervised way to learn binary target codes by minimizing the quantization loss and constraining the rotation representation invariant.
AdaLabelHash \cite{yang2021learning} proposed a binary hash function learning method with trainable variables to learn the label representation and sample representation simultaneously. 
{Similarly, the Deep Collaborative Embedding (DCE) framework \cite{li2018deep} learns target codes by integrating visual and textual information from social interactions, enhancing social image understanding through label learning in a collaborative embedding space. Additionally, the Semi-supervised Local Feature Selection (S2LFS) method \cite{li2021semi} learns target codes by selecting class-specific feature subsets, improving data classification through the combined use of labeled and unlabeled data in a semi-supervised learning framework.}

Evidently, the current target coding methods aim to enhance representation learning by creating effective target codes that directly constrain model training on a given task.
Our approach is inspired by existing target coding methods, but differs in that our proposed target codes leverage learnable parameters as a form of supervised information during training, resulting in more flexible and discriminative representations.
Additionally, our approach is the first to treat the target coding scheme as an auxiliary regularization for the classification task, thereby enhancing its effectiveness in improving deep representation learning and boosting model performance.
\vspace{-3mm}


\section{Methodology}\label{sec:methods}
In the supervised classification of images, 
given a set of training samples $\{(x_i, y_i)\}_{i=1}^N$, the deep models can be typically decomposed into a cascade of one feature representation learning module $\Phi_\text{f}$ and one classification module $\Phi_\text{c}$, which is given as
\begin{equation}
\Phi(x) = \Phi_\text{c}(\bm{z}) \circ \Phi_\text{f}(x),
\end{equation} 
where the feature representation output $\bm{z} \in \mathcal{Z}^d$ is generated from $\Phi_\text{f}(x)$, and $d$ represents the feature vector's dimension.
With the predicted output logits of different input images, the standard cross-entropy loss can be written as
\begin{equation}
\label{eq:celoss}
    L_\text{CE} = -\frac{1}{N} \sum_{i=1}^N \sum_{k=1}^K y_{i,k} \log p_{i,k},
\end{equation}
where $p_{i,k}$ is the $k$-th element of the category prediction $p_i$, $y_{i,k}$ is the corresponding label of input $x_i$, and $K$ is the number of categories. 
Inspired by the recent success of self-supervised representation learning such as 
the DTRG \cite{liu2022convolutional},
an alternative target coding scheme can be viewed as a self-supervised regularization $L_\text{reg}$ of representation learning $\Phi_\text{f}(x)$ complementary to the existing targets (\eg the one-hot codes). 
In this paper, we explore informative target coding as an auxiliary regularization for enhancing semantic representation learning.

\vspace{-2mm}
\subsection{Regularization with Hadamard Target Codes}

To improve the learning of discriminative representations, we firstly employ hand-crafted target codes 
as feature regularization for the training of deep model $\Phi(x)$. 
Specifically, the pioneering work \cite{Yang2015DeepRL} proposed a $target \ code \ \mathcal{S}$ based on the Hadamard matrix for supervising representation learning of semantic classification to discover latent dependency across classes.
The target code $\mathcal{S} \in \mathcal{T}^{K \times L}$ contains $K \ codewords$, where each $codeword$ corresponding to one semantic class consists of $L$ values from a binary alphabet set $\mathcal{T}$, \eg  $\mathcal{T}=\{1, -1\}$.
To make sure the target code has constant pairwise Hamming distance among different $codewords$ (\ie favoring for equal inter-class distance in target space), as presented in \cite{Yang2015DeepRL, yuan2020central}, the target code $\mathcal{S}$ is constructed from a Hadamard matrix $\mathcal{H} \in {\{1, -1\}}^{m \times m}$, where $m$ is a power of 2, \ie $2^n, n=1,2,\ldots,M$. 
A general method for building the Hadamard matrix is first introduced in \cite{Hedayat1978HadamardMA}, where a new Hadamard matrix can be produced by the Kronecker product of two basic Hadamard matrices as $\mathcal{H}_{2^n} = \mathcal{H}_{2^{n-1}} \otimes \mathcal{H}_2$. 
For example, the Hadamard matrix $\mathcal{H}_8$ can be produced by $\mathcal{H}_8 = \mathcal{H}_4 \otimes \mathcal{H}_2$.
Following the same setting as \cite{yuan2020central}, the target code $\mathcal{S} \in \mathcal{T}^{K \times L}$ can be randomly selected from 
the binary matrix ${\{1, -1\}}^{(m-1) \times m}$ by removing the 
first row of the Hadamard matrix $\mathcal{H}_m$, where $K \leqslant m$  and $L = m$.
Evidently, the generated target code has the following properties \cite{yuan2020central}: 1) each row of the target code has $\frac{m}{2}$ symbols that equals one; and 2) the pairwise Hamming distance is also equal to $\frac{m}{2}$ constantly.

\begin{figure}
    \centering
    \includegraphics[width=0.7\linewidth]{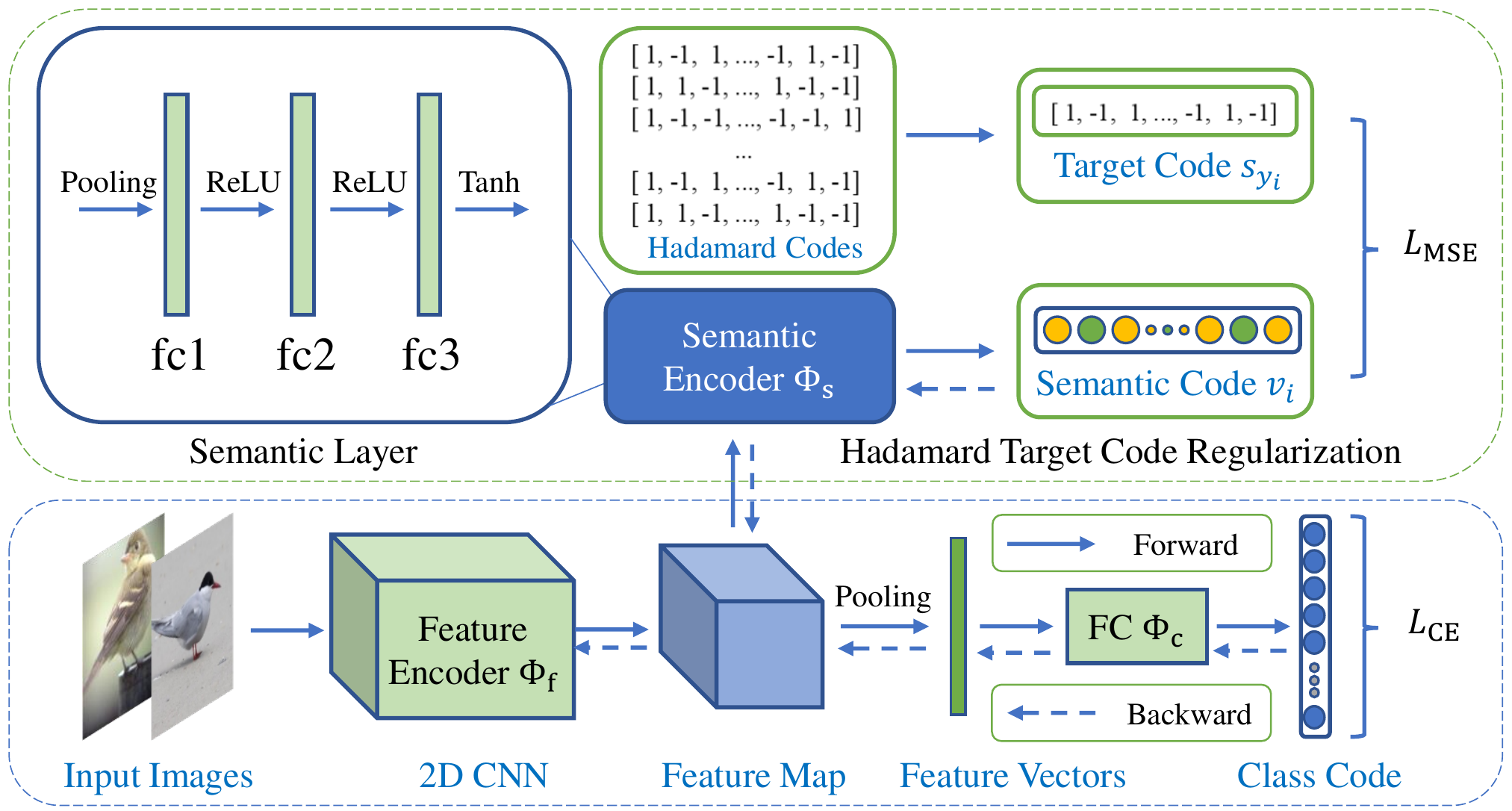}
    \caption{The pipeline of regularization with auxiliary Hadamard target codes. The lower is a cascade of representation learning and classification, while the upper is the HTC regularization module, which is not limited to specific classification networks and can be readily applied to other classifiers.
The semantic encoder aims to incorporate latent correlation across classes by the Hadamard codes into semantic representations.
    }\vspace{-2mm}
    \label{fig:framework1}
\end{figure}

For utilizing the Hadamard codes as an auxiliary regularization to supervise model training, we follow the steps below.
Firstly, we construct a $L$-dimensional Hadamard matrix as aforementioned, and randomly sample $K$ target code vectors $\bm{s}_{k} \in \mathcal{T}^{1\times L}, k=1,2,\ldots, K$, each representing one semantic category, to form a target code matrix $\mathcal{S}$.
Secondly, we adopt a semantic encoder $\Phi_\text{s}$ to encode the global feature $\bm{z}_i$ of sample $x_i$ into a semantic code vector $\bm{v}_i$, which has the same length with the target code vector $\bm{s}_{y_i} $.
The detail of semantic encoder $\Phi_\text{s}$ is illustrated in \cref{fig:framework1}, which contains three FC layers and two ReLU modules, and ends with a tanh function for converting values of the semantic code within $[-1, 1]$.
Thirdly, to make the training more efficient, we use the MSE loss as \cite{Yang2015DeepRL} to constrain training of the semantic encoder as
\begin{equation}
\label{eq:mse_loss}
    L_\text{MSE} = \frac{1}{N L} \sum_{i=1}^{N} \sum_{l=1}^L (\bm{v}_{i,l} - \bm{s}_{y_i, l} )^2,
\end{equation}
where $\bm{v}_i = \Phi_\text{s}(\bm{z}_i)$ is the predicted code vector, $\bm{s}_{y_i}$ denotes the corresponding target code vector of $x_i$, {and ``$l$" is the number of different elements in a code vector}.

The framework is illustrated as \cref{fig:framework1}, and the whole loss function for regularizing representation learning with the Hadamard Target Codes (HTC) constraint is
\begin{equation}\label{eq:hadamard_codes}
L_\text{HTC} = L_\text{CE} + \gamma L_\text{MSE},
\end{equation}
where $\gamma$ is a hyper-parameter for controlling the influence of the HTC constraint.

\begin{figure}
    \centering
    \includegraphics[width=0.7\linewidth]{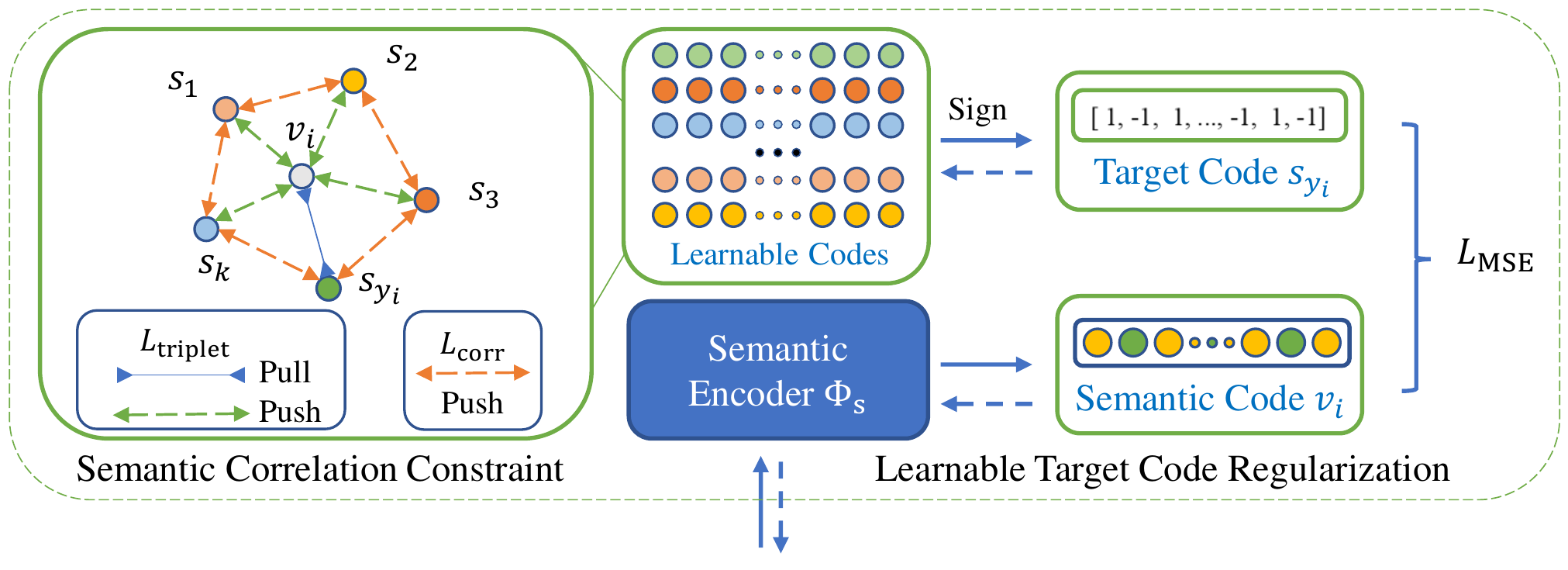}
    \caption{The module of our proposed LTC regularizer, which can replace the pre-defined Hadamard target codes with learnable parameters. 
    For imposing geometric properties, the learnable codes can be further constrained with a margin-based triplet loss for pulling samples from the same category closer and pushing samples from different categories farther. Moreover, a correlation consistency loss can ensure that the learnable target codes of different categories are orthogonally optimized with consistent semantic correlation. The semantic encoder and the main branch are similar to those shown in \cref{fig:framework1}.}\vspace{-2mm}
    \label{fig:framework2}
\end{figure}
\subsection{Learnable Target Codes}\label{subsec:LTC}

Due to the randomness introduced by the Hadamard codes, the qualities of target codes 
are usually conditioned on their length; the scheme can be less effective when the size of object categories is large while the code length is limited. 
Moreover, there could cause great computational costs when the target code from the Hadamard matrix is too long.
Therefore, we consider leveraging a learnable parameter matrix $W \in \mathbb{R}^{K \times L}$ instead of the randomly selected Hadamard codes, 
which is inspired by the AdaLabelHash \cite{yang2021learning}.
We set the parameter matrix $W$ with random initial values for the learnable target codes, where each target code $\bm{w}_k$ is a $L$-dimensional vector.

To enforce the learnable target codes into a binary vector, we adopt the sign function, \ie $\text{sgn}(\bm{w}_k)$, for converting the corresponding target code during every iteration. 
Considering the non-differentiable nature of the sign function in the chain rule, 
{we first approximate the gradient similar to straight through estimator (STE) \cite{bengio2013estimating}, and then constraint the gradient to +1 or -1 as introduced in Binarized Neural Networks (BNNs) \cite{courbariaux2016binarized}.}
For a parametric target code $\bm{w}_k$, the derivative approximation of the sign function $\text{sgn}(\bm{w}_k)$ is formulated as
\begin{equation}\label{eq:sign-derivative}
    \nabla\cdot\frac{\partial \text{sgn}(\bm{w}_k)}{\partial \bm{w}_k} = \text{min}(\text{max}(\nabla, -1), 1),
\end{equation}
{where $\nabla := \partial L /\ \partial \mathop{\mathrm{sgn}}(\bm{w}_k)$ represents the backward gradient in the chain rule.} Then, the parameter learning for target code $\bm{w}_k$ is depicted as
\begin{equation}
    \bm{w}_{k+1} \leftarrow \bm{w}_k - \eta \cdot \nabla\cdot\frac{\partial \text{sgn}(\bm{w}_k)}{\partial \bm{w}_k},
\end{equation}
where $\eta$ is the learning rate for the learnable target codes.
{For better comparison, we have also conducted experiments by adopting the tanh function with a scale  $\xi$, \ie $\tanh(\xi \cdot \bm{w}_k)$, as a differentiable surrogate of the sign function for model training.}
In our experiments (see comparative results in \Cref{table:Ablation-activation} {under default experimental settings}), the sign function we employed can consistently help the model learn better representations than the tanh function. 

Owing to the introduction of multiple positive elements in the learnable target code vector as supervision signals, inter-class correlation can be discovered and exploited to mitigate the suffering from imbalanced data, which can share similar scripts with {the DTRG \cite{liu2022convolutional} and the CSQ \cite{yuan2020central}.}

\subsection{Imposing Geometric Properties of Target Codes}                              
\label{subsec:semantic-correlation}

With the LTC as auxiliary supervision signals of deep representation learning, the key idea is how to impose desirable geometric properties of target codes across classes, which can thus enforce deep representations to favor such structural correlation.  

As illustrated in \cref{fig:framework2}, except for the MSE loss on $\bm{v}_i$ as the output of the semantic encoder module and target codes $\bm{s}_{y_i}$, the semantic feature encoding can be further constrained by a global margin-based triplet loss, which is adopted for enlarging inter-class dissimilarity and narrowing intra-class similarity in the representation space.
Technically, a semantic code $\bm{v}_i$ with a corresponding category target code $\bm{s}_{y_i}$ and $K-1$ other categories target codes $\{\bm{s}_k\}_{k \neq y_i}^K$ can construct $K-1$ triplets $(\bm{v}_i, \bm{s}_{y_i}, \bm{s}_k)$, where $k \in \{K\}_{y_j \neq y_i}$.
Before introducing the triplet loss, we would first define a semantic correlation between target code and semantic code as $\text{corr}_s(\bm{v}_i, \bm{s}_k)={\bm{v}_i}^T \bm{s}_k$. 
When $\bm{v}_i = \bm{s}_k$, the $\text{corr}_s(\bm{v}_i, \bm{s}_k)$ would get the largest value of $L$, which is the length of the code vector, while the $\text{corr}_s(\bm{v}_i, \bm{s}_k)$ would get the smallest value of $-L$ when $\bm{v}_i = -\bm{s}_k$.
With the defined semantic correlation, the loss for the global triplet pairs $(\bm{v}_i, \bm{s}_{y_i}, \bm{s}_k)$ is written as
\begin{equation}
\label{eq:triplet_loss}
L_\text{triplet} =  \frac{1}{N(K-1)} \sum_{i=1}^{N} \sum_{k \neq y_i}^K \text{max}({\bm{v}_i}^T \bm{s}_k - {\bm{v}_i}^T \bm{s}_{y_i} + \epsilon, \ 0),
\end{equation}
where $\epsilon$ is the margin parameter for constraining the correlation between positive and negative pairs.
In our LTC method, the triplet loss is employed in a global manner, removing the requirement for searching positive and negative anchors' pairs. This enables faster training and more efficient utilization of computational resources.

To improve the separability of representations, we design a correlation consistency loss to constrain different categories' target codes to have an equal semantic correlation among each other, \ie to be orthogonal.
As mentioned above, the semantic correlation among different target codes can be constructed as $\text{corr}_s(\bm{s}_k, \bm{s}_j)={\bm{s}_k}^T \bm{s}_j$.
Then, the correlation consistency loss with learnable target codes is given as
\begin{equation}
\label{eq:correlation_loss}
L_\text{corr} = \frac{1}{K (K-1)} \sum_{k=1}^{K} \sum_{j \neq k}^K |{\bm{s}_k}^T \bm{s}_j|,
\end{equation}                              
where we adopt the absolute value of the semantic correlation to enforce the correlation among different category target codes to approach zero.

\subsection{Model Training and Inference}
The overall training objective contains the ordinary cross-entropy loss on one-hot target codes in \cref {eq:celoss}, the typical MSE loss on minimizing the Euclidean distance between learnable target codes and semantic codes as \cref {eq:mse_loss}, the margin-based triplet loss in \cref {eq:triplet_loss} and the target correlation consistency loss in \cref {eq:correlation_loss}, leading to deep representation learning with an auxiliary target coding regularization as 
\begin{equation} \label{eq:whole_loss}
L_\text{LTC} = L_\text{CE} + \gamma L_\text{MSE} + \lambda L_\text{triplet} + \beta L_\text{corr},
\end{equation}
where $\lambda$ and $\beta $ are the hyper-parameters for controlling the influence of the margin-based triplet loss and the correlation consistency loss, respectively.
It is worth mentioning that, the $ L_\text{triplet}$ and the $L_\text{corr}$ are just used for constraining the learnable target codes, while the overall training objective for the HTC is illustrated in \cref {eq:hadamard_codes}.
Once trained, our model generates class predictions solely using the feature extractor $\Phi_\text{f}$ and the classifier $\Phi_\text{c}$, without the need for such regularization.
{For better understanding, we show the training and testing procedure in Algorithm \textcolor{blue}{\ref{alg:training}}.}

\begin{figure}  
\begin{center}  
\resizebox{0.85\columnwidth}{!}{
\begin{minipage}{.95\linewidth}  
\begin{algorithm}[H]
    \caption{{The pipeline of model training and testing processing}}
    \label{alg:training}
    \begin{algorithmic}[1]
    \REQUIRE The datasets $\mathcal{D}_{train}=\{(x_i, y_i)\}_{i=1}^N$ and $\mathcal{D}_{test}$, the feature encoder $\Phi_\text{f}$, the classifier $\Phi_\text{c}$, the semantic encoder $\Phi_\text{s}$, the class number ${K}$, the code length ${L}$.
    \ENSURE The predicted output $\bm{p}_i$ and the learned parameters

		    \IF {regularizing with LTC}
	    		\STATE Random initialize the Learnable target code matrix $W \in \mathbb{R}^{K \times L}$
	    	\ELSIF {regularizing with HTC}
	    		\STATE Random select the Hadamard target code matrix $\mathcal{S} \in \mathbb{R}^{K \times L}$
           	\ENDIF 
	    	    
    \IF{training}
            \STATE Random sample a batch of training data $\mathcal{B} \subset \mathcal{D}_{train}$
            \FOR{$i = 1$ \textbf{to} $|\mathcal{B}|$}
            	\STATE Obtain the feature vector ${\bm{z}}_i = \Phi_\text{f}({\bm{x}}_i)$
            	\STATE Obtain the prediction output ${\bm{p}}_i = \Phi_\text{c}({\bm{z}}_i)$
            	\IF {regularizing with LTC or HTC}
	                \STATE Obtain the semantic code vector $\bm{v}_i = \Phi_\text{s}(\bm{z}_i)$ 
            	\ENDIF
            \ENDFOR
            \IF {regularizing with LTC}
               	\STATE Compute total loss according with $L_\text{LTC}$ in \textbf{Eq.} (\ref{eq:whole_loss})
            	\STATE Backward to update the parameters of $\Phi_\text{f}$, $\Phi_\text{s}$ and $\Phi_\text{c}$, and the matrix $W$
           	\ELSIF {regularizing with HTC}
           		\STATE Compute total loss according with $L_\text{HTC}$ in \textbf{Eq.} (\ref{eq:hadamard_codes})            	
           		
	            \STATE Backward to update the parameters of $\Phi_\text{f}$, $\Phi_\text{s}$ and $\Phi_\text{c}$    
            \ELSE
	            \STATE Compute total loss according with $L_\text{CE}$ in \textbf{Eq.} (\ref{eq:celoss})                
	            \STATE Backward to update the parameters of $\Phi_\text{f}$ and $\Phi_\text{c}$   
            \ENDIF  
    \ELSE
        \STATE Random sampling testing data $x_i \subset \mathcal{D}_{test}$    
        \STATE Compute the predicted output $\bm{p}_i = \Phi_\text{c}(\bm{z}) \circ \Phi_\text{f}(x_i)$
    \ENDIF
    \RETURN the predicted output $\bm{p}_i$ and the learned parameters

\end{algorithmic}
\end{algorithm}
\end{minipage}}
\end{center}
\end{figure}


\section{Experiments}\label{sec:exps}

In this section, we first introduce the details of datasets and experimental settings.
Then, we empirically validate the effectiveness of the HTC and our proposed LTC methods on three popular benchmarks of fine-grained image classification. 
{Next, the proposed LTC method is applied to the fine-grained retrieval task to verify its effectiveness on deep metric learning.}
Additionally, the proposed method is further adopted for three data-imbalanced learning tasks for verifying the superior robustness. 
Finally, the projected representations based on the t-SNE \cite{Maaten2008VisualizingDU} and the learned semantic correlations between different categories are visualized for a better understanding.

\subsection{Experimental Details} \label{subsec:exp-detials}
\begin{figure}[t]
    \centering
    \includegraphics[width=0.7\linewidth]{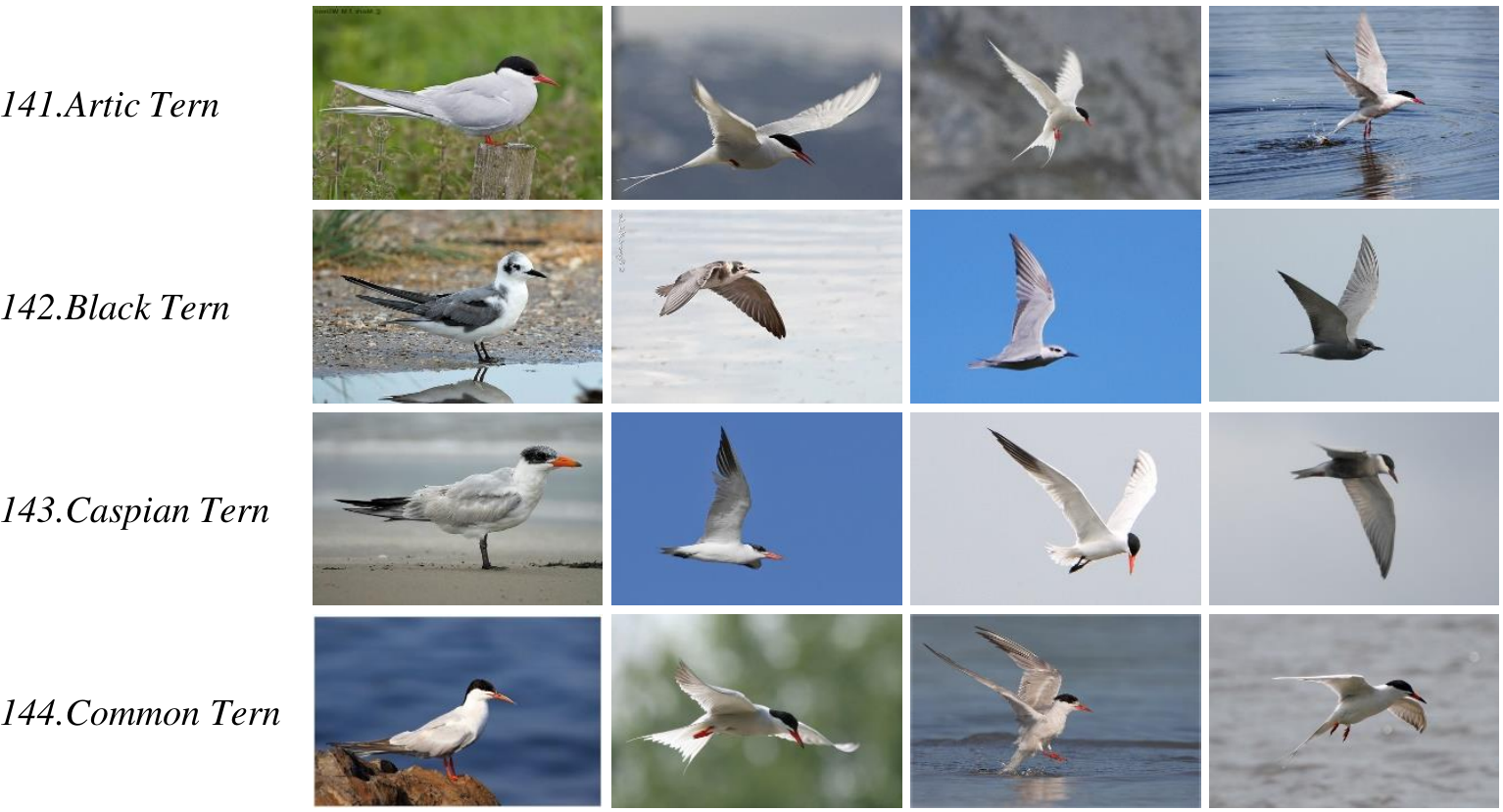}
    \caption{Examples of four different Tern species from the CUB \cite{wah2011caltech}. One species in each row is with different instances. We can find that there are usually existing small inter-class variations among different fine-grained species and large intra-class variations in the same class.
    }
    \label{fig:cub_examples}
\end{figure}

\noindent\textbf{Datasets --}
In the problem of representation learning in fine-grained visual classification, We firstly validate our proposed target codes constraint methods on three popular fine-grained benchmarks, \ie the CUB-200-2011 (CUB) \cite{wah2011caltech}, the Stanford Cars (CAR) \cite{krause20133d}, and the FGVC Aircraft (AIR) \cite{maji2013fine}. 
Then, we conduct experiments on the following commonly used long-tailed datasets: the CIFAR100-LT \cite{cao2019learning} and the ImageNet-LT \cite{liu2019large}, for evaluating the effectiveness of imbalanced data learning. 
The details of the three fine-grained and the three long-tailed datasets are provided in \Cref{table:datasets}, where the last column presents the imbalance ratio of different datasets.
In general, the data splits of the three fine-grained datasets are basically uniformly distributed among different categories, by following that provided in \cite{liu2022convolutional}.
Meanwhile, the imbalanced CIFAR-100-LT and ImageNet-LT datasets are generated from the vanilla CIFAR \cite{krizhevsky2009learning} and ImageNet \cite{deng2009imagenet} datasets for constructing imbalances.
Moreover, the iNaturalist18 \cite{Horn2018TheIS} dataset is another extremely imbalanced and large-scale dataset with a large number of categories at the same time.
In all our experiments, only the category labels are used as the supervision information for model training without adopting any additional prior information.
To illustrate the challenge of representation learning in fine-grained visual analysis, the selected example images are visualized in Fig. \ref{fig:cub_examples}.

\begin{table}[t]
\centering
\caption{The statistics of classification datasets. The imbalance ratio is defined by the $\text{max}_k (n_k)/\text{min}_k (n_k)$, where $n_k$ is the number of samples from the k-th class.} \label{table:datasets}
\resizebox{0.6\columnwidth}{!}{
\begin{tabular}{@{}lcccc@{}}
\toprule
Dataset                             & \# Class  & \# Train    & \# Test  	 & \# Imbalance    \\ \midrule
CUB \cite{wah2011caltech}           & 200      & 5,994            & 5,794           & 1               \\
CAR \cite{krause20133d}             & 196      & 8,144            & 8,041           & 1               \\
AIR \cite{maji2013fine}             & 100      & 6,667            & 3,333           & 1               \\ \midrule
CIFAR100-LT \cite{cao2019learning}   & 100      & 50K             & 10K          & \{10, 50, 100\} \\
ImageNet-LT \cite{liu2019large}      & 1,000    & 115.8K          & 50K          & 256    \\
iNaturalist18 \cite{Horn2018TheIS}   & 8,142    & 437.5K          & 24K          & 500             \\  \bottomrule
\end{tabular}}
\end{table}

\vspace{0.1cm}\noindent\textbf{Comparative Methods --}
To evaluate the effectiveness of our method on fine-grained image classification, we first conduct experiments by employing three variants of the ResNet \cite{He2016DeepRL}, \ie ResNet18, ResNet34, and ResNet50, which are all pre-trained on the ImageNet \cite{deng2009imagenet} dataset. We compare our proposed LTC with the baseline, the Label Smoothing \cite{Szegedy2016RethinkingTI}, the Online Label Smoothing \cite{zhang2020delving}, the Center Loss \cite{wen2016discriminative}, the DTRG \cite{liu2022convolutional} and the HTC on the CUB, the CAR, and the AIR datasets, respectively. 
{And, we also compare the proposed LTC method with some related transformer-based methods on the CUB dataset.
Then, we conduct experiments on the fine-grained image retrieval task {with the BN-Inception \cite{Ioffe2015BatchNA}} by following the setting of Circle Loss \cite{sun2020circle} and Multi-similarity Loss (MS) \cite{wang2019multi} on the CUB \cite{wah2011caltech} and the CAR \cite{krause20133d} datasets.}
On the evaluation of the robustness against imbalanced data, we adopt ResNet32 \cite{He2016DeepRL} as the backbone for CIFAR100-LT \cite{cao2019learning}, and ResNet50 \cite{He2016DeepRL} as the backbone for the ImageNet-LT \cite{liu2019large} and the Naturalist18 \cite{Horn2018TheIS} like most of previous methods. 
Except for comparison with the baseline, we also compare our method with several popular related methods, \eg CB \cite{Cui2019ClassBalancedLB}, LDAM \cite{Cao2019LearningID}, IB Loss \cite{Park2021InfluenceBalancedLF}, and Decouple-based methods \cite{Kang2020Decoupling}. 
It is important to note that all models for imbalanced data learning are trained from scratch.

\vspace{0.1cm}\noindent\textbf{Implementation Details --}
In the experiments for fine-grained classification, we adopt stochastic gradient descent (SGD) with a momentum of 0.9 and a weight decay of $1\times10^{-4}$ for CNN-based model training. Following the implementation details in \cite{liu2022convolutional}, we train these experiments for 100 epochs with a batch size of 16 and step-wise learning rate decay.
Since the backbones are pre-trained on the ImageNet, the initial learning rate is set as 0.001 for the feature extractor and 0.01 for new parameters, which contain the classifier and the semantic encoder.
{The images are first resized to $512 \times 512$, and then cropped with size $448 \times 448$ for batch training and testing.} 
For the learnable target codes, the initial learning rate is set as 0.1.
Then, the initial learning rate is decayed with a factor of 0.1 at the 40-th epoch and the 70-th epoch, respectively.  
In \cref {eq:whole_loss}, there are three hyperparameters, \ie $\gamma$ for controlling semantic codes learning, $\lambda$ for controlling category relation learning, and $\beta$ for constraining the correlation consistency. 
{Unless particularly stated, we set $\gamma=1$, $\lambda=0.01$, and $\beta=0.1$ in our experiments by the ablation studies.}
The length of the target code is set as 512 by our ablation studies, and the margin parameter $\epsilon$ in \cref {eq:correlation_loss} is usually suggested to be set as equal to the length of the target code. 
{For transformer-based model training, we follow the setting from TransFG \cite{he2022transfg} with the same hyper-parameters as introduced before.}

{In the experiments for fine-grained retrieval, we follow the same setting from Multi-similarity Loss (MS) \cite{wang2019multi}, where the model used the first half of the classes as the training set and the second half of the classes as the test set. Considering the small pair-wise loss in the MS method, we set the $\gamma=0.2$, $\lambda=0.001$, and $\beta=0.01$ for better convergence, and the initial learning rate as $2e-4$ for the learnable target codes.}

In the experiments for imbalanced data classification, we prepare the CIFAR100-LT dataset and implement the model training by following \cite{cao2019learning, Park2021InfluenceBalancedLF}. 
The model is trained from scratch for 200 epochs with an initial learning rate of 0.1, which is decayed by 0.1 at the 160-th epoch and the 180-th epoch, respectively. The hyper-parameters for our target code constraint are set as same as that introduced before, except for $\epsilon=L/2$.
{For the ImageNet-LT and the Naturalist18 datasets, the images are both cropped to size $224 \times 224$.}
In the experiments on the ImageNet-LT, we train the model from scratch for 100 epochs with a batch size of 128, an initial learning rate of 0.1, and a cosine learning rate schedule for better optimization. 
Except for $\lambda=0.005$ and $\beta=0.05$, other hyper-parameters are set as same as that introduced before.
While on the Naturalist18, we train the model similarly to the ImageNet-LT except with 200 training epochs.
{We implement and train the models in all experiments with the Pytorch library on a single NVIDIA Tesla V100 GPU and Intel Xeon CPU running at 2.2 GHz.
}

\begin{table}[t]
\centering
\caption{Comparison with classification accuracy using ResNet18, ResNet34, and ResNet50 backbones on the CUB, the CAR, and the AIR datasets \protect\footnotemark[2].} 
\label{table:fine-grained-data}
\resizebox{0.6\columnwidth}{!}{
\begin{tabular}{@{}lcccc@{}}
\toprule
Method                                          & Backbone        & CUB (\%)       & CAR (\%)       & AIR (\%)       \\ \midrule
Baseline                                        & ResNet18        & 81.88          & 90.45          & 88.60          \\
LS \cite{Szegedy2016RethinkingTI}               & ResNet18        & 83.59          & 91.22          & 87.70          \\
OLS \cite{zhang2020delving}                     & ResNet18        & 83.52          & 91.13          & 87.46          \\
Center Loss \cite{wen2016discriminative}        & ResNet18        & {\ul 84.12}    & {\ul 92.53}    & 89.23          \\
DTRG    \cite{liu2022convolutional}             & ResNet18        & \textbf{84.52} & 91.99          & {\ul 90.28}    \\ 
HTC                                             & ResNet18        & 83.19          & 91.54          & 88.93          \\ 
LTC (Ours)                                      & ResNet18        & {84.07}        & \textbf{92.67} & \textbf{90.91} \\ \midrule \midrule
Baseline                                        & ResNet34        & 84.82          & 91.94          & 89.98          \\
LS \cite{Szegedy2016RethinkingTI}               & ResNet34        & 85.16          & 92.81          & 91.18          \\
OLS \cite{zhang2020delving}                     & ResNet34        & 85.26          & 92.60          & 90.25          \\
Center Loss \cite{wen2016discriminative}        & ResNet34        & \textbf{85.76} & {92.76}        & 91.21          \\
DTRG    \cite{liu2022convolutional}             & ResNet34        & {\ul 85.66}    & {\ul 93.32}    & {\ul 91.54}    \\ 
HTC                                             & ResNet34        & 85.35          & 92.97          & 91.33          \\ 
LTC (Ours)                                      & ResNet34        & {85.55}        & \textbf{94.08} & \textbf{92.35} \\ \midrule \midrule
Baseline                                        & ResNet50        & 85.46          & 92.89          & 90.97          \\
LS \cite{Szegedy2016RethinkingTI}               & ResNet50        & 86.02          & 93.41          & 92.00          \\
OLS \cite{zhang2020delving}                     & ResNet50        & 86.32          & 93.31          & 91.21          \\ 
Center Loss \cite{wen2016discriminative}        & ResNet50        & 86.41          & 93.52          & 91.40          \\
DTRG    \cite{liu2022convolutional}             & ResNet50        & \textbf{87.29} & \textbf{94.31} & {\ul 92.12}    \\ 
HTC                                             & ResNet50        & 86.33          & 93.92          & 91.36          \\
LTC (Ours)                                      & ResNet50        & {\ul 86.90}    & {\ul 94.27}    & \textbf{92.77} \\ \bottomrule
\end{tabular}}
\end{table}

\subsection{Evaluation on Fine-grained Classification}
\noindent
\textbf{Results on CNN-based models.}
As mentioned in \cref {subsec:exp-detials}, we first conduct experiments on the CUB, the CAR, and the AIR datasets with ResNet18, ResNet34, and ResNet50 backbone networks, respectively. 
Comparative results are reported in \Cref{table:fine-grained-data}, where all methods are under identical experimental settings.
Evidently, the proposed LTC can consistently achieve performance improvements over the baseline and the HTC.
Meanwhile, our proposed LTC can always achieve competitive results than the Label Smoothing (LS) \cite{Szegedy2016RethinkingTI}, the Online Label Smoothing (OLS) \cite{zhang2020delving}, the Center Loss \cite{wen2016discriminative}, and even the recent state-of-the-art method DTRG \cite{liu2022convolutional}, which further verifies the effectiveness of our proposed method.
{Specifically, on the AIR dataset, our proposed LTC method consistently outperforms the HTC method by approximately 1\% and outperforms the baseline by more than 2\%.}
The experimental results reveal that both the HTC and the LTC regularization can effectively improve the performance of the deep models on fine-grained image classification tasks. 
{Additionally, we conduct more comparative experiments on several other different CNN backbones, \ie VGG16 \cite{simonyan2014very}, BN-Inception \cite{Ioffe2015BatchNA}, MobileNetV2 \cite{sandler2018mobilenetv2}, and DenseNet121 \cite{huang2017densely}. As reported in Table \ref{table:backbones},}
the LTC can consistently achieve even better results than the HTC, indicating that the flexibility of the learnable codes and their ability to reveal the geometric structure of the label space are advantageous for representation learning. \textcolor{black}{We believe that the observed improvement is significant and indicative of the potential usefulness of our proposed method in practical applications.}
\footnotetext[2]{All reported results are implemented in the same setting. The best results are marked in bold, and the second-best results are marked with underlined.}

\begin{table}[htp]
\centering
\caption{{Comparison of classification accuracy using VGG16, BN-Inception, MobileNetV2, and DenseNet121 backbones on the CUB, the CAR, and the AIR datasets \protect\footnotemark[2].}}\label{table:backbones}
\setlength{\tabcolsep}{3.mm}{
\resizebox{0.6\columnwidth}{!}{
\begin{tabular}{@{}lcccc@{}}
\toprule
Backbone                      & Method   & CUB (\%)       & CAR (\%)       & AIR (\%)       \\ \midrule
\multirow{3}{*}{VGG16 \cite{simonyan2014very}}         & Baseline & 78.56          & 87.30          & 86.17          \\
                              & HTC      & 80.22          & 88.57          & 87.25          \\
                              & LTC      & \textbf{80.76} & \textbf{89.44} & \textbf{88.78} \\ \midrule
\multirow{3}{*}{BN-Inception \cite{Ioffe2015BatchNA}}  & Baseline & 82.53          & 91.16          & 89.80          \\
                              & HTC      & 83.43          & 92.50          & 91.03          \\
                              & LTC      & \textbf{83.60} & \textbf{92.87} & \textbf{91.33} \\ \midrule
\multirow{3}{*}{MobileNetV2 \cite{sandler2018mobilenetv2}}     & Baseline & 82.45          & 90.66          & 88.54          \\
                              & HTC      & 83.66          & 91.89          & 90.22          \\
                              & LTC      & \textbf{83.83} & \textbf{92.64} & \textbf{91.69} \\ \midrule
\multirow{3}{*}{DenseNet121 \cite{huang2017densely}}   & Baseline & 85.55          & 92.21          & 91.57          \\
                              & HTC      & 86.11          & 93.23          & 92.41          \\
                              & LTC      & \textbf{86.26} & \textbf{93.96} & \textbf{93.16} \\ \bottomrule
\end{tabular}}}
\end{table}

\noindent
{\textbf{Results on Transformer-based models.}
Given the recent prominence of transformer models, it is essential to compare our approach with transformer-based methods to demonstrate its effectiveness and superiority. Accordingly, we extend our comparison with the latest transformer-based methods such as TransFG \cite{he2022transfg}, RAMS-Trans \cite{hu2021rams}, and AFTrans \cite{zhang2022free}, which are evaluated on the CUB dataset used in our experiments. The comparison results are presented in \Cref{table:transformer}. We have carefully evaluated the experimental results to ensure that our method achieves competitive or superior performance compared to the state-of-the-art transformer-based methods.}

\begin{table}[t]
\centering
\caption{{Classification {accuracy} comparison of different transformer-based methods on the CUB dataset \protect\footnotemark[3].}}  
\label{table:transformer}
\resizebox{0.4\columnwidth}{!}{
\begin{tabular}{@{}lcc@{}}
\toprule
Method      & Backbone   & CUB (\%)  \\ \midrule
ViT \cite{Dosovitskiy2021AnII}$^{\ddagger}$          & ViT-B\_16 & 90.3 \\
ViT+LTC (Ours)     & ViT-B\_16 & 90.6 \\ \midrule
TransFG \cite{he2022transfg}$^{\ddagger}$      & ViT-B\_16 & 91.0   \\
RAMS-Trans \cite{hu2021rams}$^{\dagger}$   & ViT-B\_16 & 91.3   \\
AFTrans \cite{zhang2022free}$^{\dagger}$ & ViT-B\_16 & \textbf{91.5} \\
TransFG+LTC (Ours) & ViT-B\_16 & \textbf{91.5} \\ \bottomrule
\end{tabular}}
\vspace{-0.2cm}
\end{table}
\footnotetext[3]{"$\dagger$" indicates that the results are from the original paper, and "$\ddagger$" indicates our re-implemented results in the same experimental settings. The best results are marked in bold.}

\begin{table*}[t]
\centering
\caption{{Comparison of Recall@K (\%) performance on the CUB and the CAR datasets for retrieval, where superscript denotes the embedding size \protect\footnotemark[3].}}\label{table:retrieval}
\setlength{\tabcolsep}{3.mm}{
\resizebox{0.8\columnwidth}{!}{
\begin{tabular}{@{}lcccclcccc@{}}
\toprule
\multirow{2}{*}{Method} & \multicolumn{4}{c}{CUB-200-2011 (\%)}   &  & \multicolumn{4}{c}{CAR-196 (\%)}     \\ \cmidrule(lr){2-5} \cmidrule(l){7-10} 
                        									& R@1          	& R@2          	& R@4          	& R@8    &         	& R@1          & R@2          & R@4          & R@8     \\ \midrule
$\text{Margin}^{128}$ \cite{wu2017sampling}        			& 63.6   		& 74.4   		& 83.1   		& 90.0  &  			& 79.6 			& 86.5 			& 91.9 			& 95.1 \\
$\text{HTL}^{512}$ \cite{ge2018deep}$^{\dagger}$ 	 		& 57.1	 		& 68.8	 		& 78.7	 		& 86.5	 &  		& 81.4	   		& 88.0	 		& 92.7	 		& 95.7 \\
$\text{ABE}^{512}$ \cite{kim2018attention}$^{\dagger}$   	& 60.6 			& 71.5 			& 79.8 			& 87.4 &  			& \textbf{85.2}	& 90.5 			& 94.0 			& 96.1 \\                        
$\text{MS}^{512}$ \cite{wang2019multi}$^{\dagger}$           & 65.7         	& 77.0         	& \textbf{86.3} & 91.2 &        	& 84.1         	& 90.4      	& 94.0        	& 96.5 \\
$\text{CircleLoss}^{512}$ \cite{sun2020circle}$^{\dagger}$   & 66.7      	& 77.4      	& 86.2     		& 91.2 &        	& 83.4       	& 89.8         	& {94.1} 		& 96.5 \\ \midrule\midrule
$\text{Margin}^{512}$ \cite{wu2017sampling}$^{\ddagger}$  	& 65.2   		& 75.8   		& 84.1   		& 90.1  &			& 79.6 			& 87,3 			& 92.1 			& 95.5 \\
$\text{Margin+LTC}^{512}$     (Ours)           				& 66.4   		& 76.4   		& 84.3   		& 90.6  &			& 82.4 			& 89.0 			& 93.1 			& 96.0 \\ \midrule\midrule
$\text{MS}^{512}$ \cite{wang2019multi}$^{\ddagger}$ 			& 65.9			& 77.2			& 85.1			& 91.0 &  			& 82.7 			& 89.9 			& 93.9 			& 96.4 \\		
$\text{MS+LTC}^{512}$ (Ours)                    			& \textbf{67.6} & \textbf{77.9} & 85.7 			& \textbf{91.2} &  	& {84.7} 		& {90.5} 		& {93.9} 		& {96.7} \\ \midrule\midrule
$\text{CircleLoss}^{512}$ \cite{sun2020circle}$^{\ddagger}$ 	& 65.2   		& 75.6   		& 84.4   		& 90.8  &			& 83.1 			& 89.8 			& 94.0 			& 96.6 \\
$\text{CircleLoss+LTC}^{512}$  (Ours)         				& 66.5   		& 76.4   		& 84.1   		& 90.6  &			& \textbf{85.1}	& \textbf{90.9} & \textbf{94.6}	& \textbf{96.8} \\ \bottomrule
\end{tabular}}}
\end{table*}

\subsection{Evaluation on Fine-grained Retrieval}
{To further illustrate the effectiveness of our method for deep representation learning, we also conduct experiments on the image retrieval task.
We evaluate our proposed LTC method on two fine-grained image retrieval benchmarks, i.e., CUB \cite{wah2011caltech} and CAR \cite{krause20133d}. 
To make our method suitable for the fine-grained retrieval task on CUB and CAR datasets, we similarly treat our proposed LTC as the auxiliary constraint for model optimization. 
{Specifically, we just change the classification pipeline as shown in Fig. \ref{fig:framework1} to the retrieval framework for model training. The results compared with the Margin Loss \cite{wu2017sampling}, Multi-similarity Loss \cite{wang2019multi}, Circle Loss \cite{sun2020circle}, and other most related methods are presented in Table \ref{table:retrieval}.}
Absolutely, we can find that the model trained with our proposed LTC constraint can always achieve superior results at the Recall@1 values, both on the CUB and CAR datasets.}

\subsection{Evaluation on Imbalanced Data Learning}
In order to verify the robustness of our proposed method against imbalanced data distribution, we conduct experiments on the CIFAR-100-LT, the ImageNet-LT, and the Naturalist18 datasets. The imbalance ratio and statistical details of different datasets are shown in \Cref{table:datasets}. The larger the imbalance ratio, the more imbalanced the dataset.

\begin{table}[htp]
\centering
\caption{Classification accuracy (\%) of ResNet-32 on the CIFAR-100-LT datasets \protect\footnotemark[3].} \label{table:cifar-data}
\resizebox{0.48\columnwidth}{!}{
\begin{tabular}{@{}lccc@{}}
\toprule
Imbalance ratio                                                 & 100            & 50             & 10             \\ \midrule
CB \cite{Cui2019ClassBalancedLB}$^{\dagger}$                    & 39.60           & 45.32          & 57.99          \\
IB \cite{Park2021InfluenceBalancedLF}$^{\dagger}$               & {42.14} & 46.22          & 57.13          \\
IB + CB \cite{Park2021InfluenceBalancedLF}$^{\dagger}$          & 41.31          & 46.16          & 56.78          \\
IB + Focal \cite{Park2021InfluenceBalancedLF}$^{\dagger}$       & 42.06          & {47.79}        & {58.20}  \\ \midrule \midrule
CE                                                              & 38.43          & 43.66           & 56.91          \\
HTC                                                        & 39.38          & 43.97          & 57.20          \\
LTC (Ours)                                                 & \textbf{39.78} & \textbf{44.88} & \textbf{57.47} \\ \midrule \midrule
CE-DRW \cite{Cao2019LearningID}$^{\ddagger}$                    & 41.76          & 45.84          & 57.97          \\
HTC-DRW                                                    & 42.50          & 46.86          & 58.30          \\
LTC-DRW (Ours)                                               & \textbf{43.67} & \textbf{48.77} & \textbf{58.94} \\ \midrule \midrule
LDAM-DRW \cite{Cao2019LearningID}$^{\ddagger}$                  & 42.70          & 47.29          & 57.51          \\
HTC-LDAM-DRW                                                   & 42.63          & 46.96          & 57.33          \\
LTC-LDAM-DRW (Ours)                                               & \textbf{44.60} & \textbf{48.14} & \textbf{58.92} \\ \bottomrule
\end{tabular}}
\end{table}

\noindent\textbf{Results on the CIFAR-LT.}
We first conduct experiments on the CIFAR100-LT dataset with ResNet32 backbone by using different imbalance ratios: 10, 50, and 100, whereas the validation set is still balanced. 
The results are reported in \Cref{table:cifar-data}. It is observed that adopting our proposed LTC method can always achieve significant improvements on different imbalance ratio datasets, while performance achieved with the proposed HTC is slightly inferior. 
This phenomenon suggests that our proposed LTC regularization can help the model learn more class-wise relational information, which further improves representation learning.
Especially, our proposed LTC method with CE-DRW or LDAM-DRW frameworks can achieve much better performance under different imbalance ratios than existing state-of-the-art methods.
{Note that, the CE-DRW strategy and LDAM-DRW method are both proposed by Cao \etal \cite{Cao2019LearningID} for regularizing the minority classes with a deferred re-weighting (DRW) training schedule and a label-distribution-aware margin (LDAM) loss function respectively.}

\noindent\textbf{Results on the ImageNet-LT.}
To further verify the effectiveness of our method on large-scale data, another experiment on ImageNet-LT is conducted using the ResNet50 backbone.
Results of competitive methods are compared in \Cref{table:ImageNet-data}, where our proposed LTC methods can achieve superior improvements than the baseline by about \textbf{4\%} in top-1 accuracy. When adopting our LTC method with CE-DRW framework, our method can outperform the baseline by \textbf{4.38\%}, which is also superior to other existing methods.
The performance comparison between our proposed HTC and LTC can again verify that learnable target codes are much better than hand-crafted target codes for class-wise correlation learning when used as the regularization.

\begin{table}[htp]
\centering
\caption{Comprehensive classification accuracy (\%) of ResNet50 with different methods on ImageNet-LT \protect\footnotemark[3].} 
\label{table:ImageNet-data}
\resizebox{0.4\columnwidth}{!}{
\begin{tabular}{@{}lcc@{}}
\toprule
\multirow{2}{*}{Method} & \multicolumn{2}{c}{ImageNet-LT} \\ \cmidrule(l){2-3} 
                                                                & top-1 (\%)     & top-5 (\%)     \\ \midrule
Decouple-NCM \cite{Kang2020Decoupling}$^{\dagger}$              & 44.30          & -              \\
Decouple-cRT \cite{Kang2020Decoupling}$^{\dagger}$              & 47.30          & -              \\
Decouple-LWS \cite{Kang2020Decoupling}$^{\dagger}$              & 47.70          & -              \\ 
Center Loss \cite{wen2016discriminative}$^{\ddagger}$           & 43.22 	    & 68.91          \\ 
DTRG        \cite{liu2022convolutional}$^{\ddagger}$            & 43.79 	    & 69.35          \\  \midrule \midrule
CE                                                              & 41.30          & 65.98          \\
HTC                                                        & 42.28 	     & 67.28          \\
LTC (Ours)                                                   & \textbf{45.25} & \textbf{70.59} \\ \midrule \midrule
CE-DRW \cite{Cao2019LearningID}$^{\ddagger}$                    & 43.69          & 67.84          \\
HTC-DRW                                                    & 44.52          & 68.71          \\
LTC-DRW (Ours)                                              & \textbf{48.07} & \textbf{72.48} \\ \bottomrule
\end{tabular}}
\end{table}

\begin{table}[htp]
\centering
\caption{Comparison of classification accuracy (\%) on the iNaturalist18 dataset using ResNet50 backbone \protect\footnotemark[3].} 
\label{table:iNaturalist18-data}
\resizebox{0.4\columnwidth}{!}{
\begin{tabular}{@{}lcc@{}}
\toprule
\multirow{2}{*}{Method} & \multicolumn{2}{c}{iNaturalist18} \\ \cmidrule(l){2-3} 
                                                            & top-1 (\%)     & top-5   (\%)     \\ \midrule
CB \cite{Cui2019ClassBalancedLB}$^{\dagger}$                 & 61.12          & 81.03            \\
LDAM \cite{Cao2019LearningID}$^{\dagger}$                   & 64.58          & 83.52            \\
IB \cite{Park2021InfluenceBalancedLF}$^{\dagger}$           & 65.39          & 84.98            \\
CE                                                          & 64.21          & 84.35            \\
LTC (Ours)                                                  & 65.04         & 85.40        \\ \midrule \midrule
CE-DRW \cite{Cao2019LearningID}$^{\ddagger}$                & 67.03          & 84.75            \\
LDAM-DRW \cite{Cao2019LearningID}$^{\dagger}$               & 68.00          & 85.18            \\
LTC-DRW (Ours)                                       & \textbf{68.48}         & \textbf{86.26}        \\ \bottomrule
\end{tabular}}
\end{table}

\vspace{0.1cm}\noindent\textbf{Results on the iNaturalist18.}
We also evaluate our method on a real-world long-tailed dataset, iNaturalist18, with 8,142 categories as presented in \Cref{table:datasets}. Here, we only conduct experiments using the learnable target codes since Hadamard codes are limited by the large number of categories in this dataset. Comparative evaluation with state-of-the-art methods is reported in \Cref{table:iNaturalist18-data}. Although this is a large-scale dataset, our method still achieves considerable improvements in image classification with the help of auxiliary target codes learning.

\subsection{Ablation Study}
\vspace{0.1cm}\noindent\textbf{Impact of the Proposed Components.}
As illustrated in \cref {eq:whole_loss}, there are three key components in our proposed LTC regularization -- the MSE loss in \cref {eq:mse_loss} for constraining the learning of semantic encoder, the margin-based triplet loss in \cref {eq:triplet_loss} for constraining the semantic correlation among different categories, and the target codes correlation consistency loss in \cref {eq:correlation_loss} for constraining the learning of target codes. Due to the different roles played by different components, we conduct ablation studies to demonstrate the effectiveness of different components in our LTC model. 
{Experiments are conducted on three fine-grained datasets, \ie the CUB, the CAR, the AIR, and one imbalance dataset, \ie the ImageNet-LT, with ResNet18 backbone and default experimental settings for classification.}
From \Cref{table:Ablation-parameter}, we can observe that the proposed components in our LTC model are all verified to be effective in different fine-grained and imbalanced datasets. 
Especially, when all three components of the LTC model are adopted, our method consistently outperforms the baseline by more than \textbf{2\%} in terms of top-1 classification accuracy across different fine-grained datasets, {while by more than \textbf{1\%} for the ImageNet-LT dataset}. It is important to note that the three components are always used together in the LTC regularization for all other experiments.

\begin{table}[t]
\centering
\caption{Ablation studies about different losses on three fine-grained datasets and one imbalance dataset with ResNet18 backbone and default experimental settings for classification {(accuracy)}.} 
\label{table:Ablation-parameter}
\vspace{2mm}
\setlength{\tabcolsep}{2.mm}{
\resizebox{0.7\columnwidth}{!}{
\begin{tabular}{@{}l|ccc|cccc@{}}
\toprule
Method                  & $L_\text{MSE}$  & $L_\text{triplet}$    & $L_\text{corr}$   		& CUB (\%)       & CAR (\%)       & AIR (\%)    	& {ImageNet-LT} (\%)	\\ \midrule
\multirow{4}{*}{LTC} 	&                 &                       &                       	& 81.88          & 90.45          & 88.60       	&40.25    	 		\\
                          & $\surd$         &                       &                       & 82.91          & 91.48          & 88.93       	&40.62    			\\
                          & $\surd$         & $\surd$               &                       & 83.88          & 92.65          & 90.70         	&41.15 		 		\\
                          & $\surd$         & $\surd$               & $\surd$               & \textbf{84.07} & \textbf{92.67} & \textbf{90.91} 	&\textbf{41.36} 	\\ \bottomrule
\end{tabular}}}
\end{table}

\begin{figure*}[htp]
    \centering
    \subfigure[Impact of $\gamma$]{ 
            \includegraphics[width=0.31\linewidth, trim=5 0 60 50,clip]{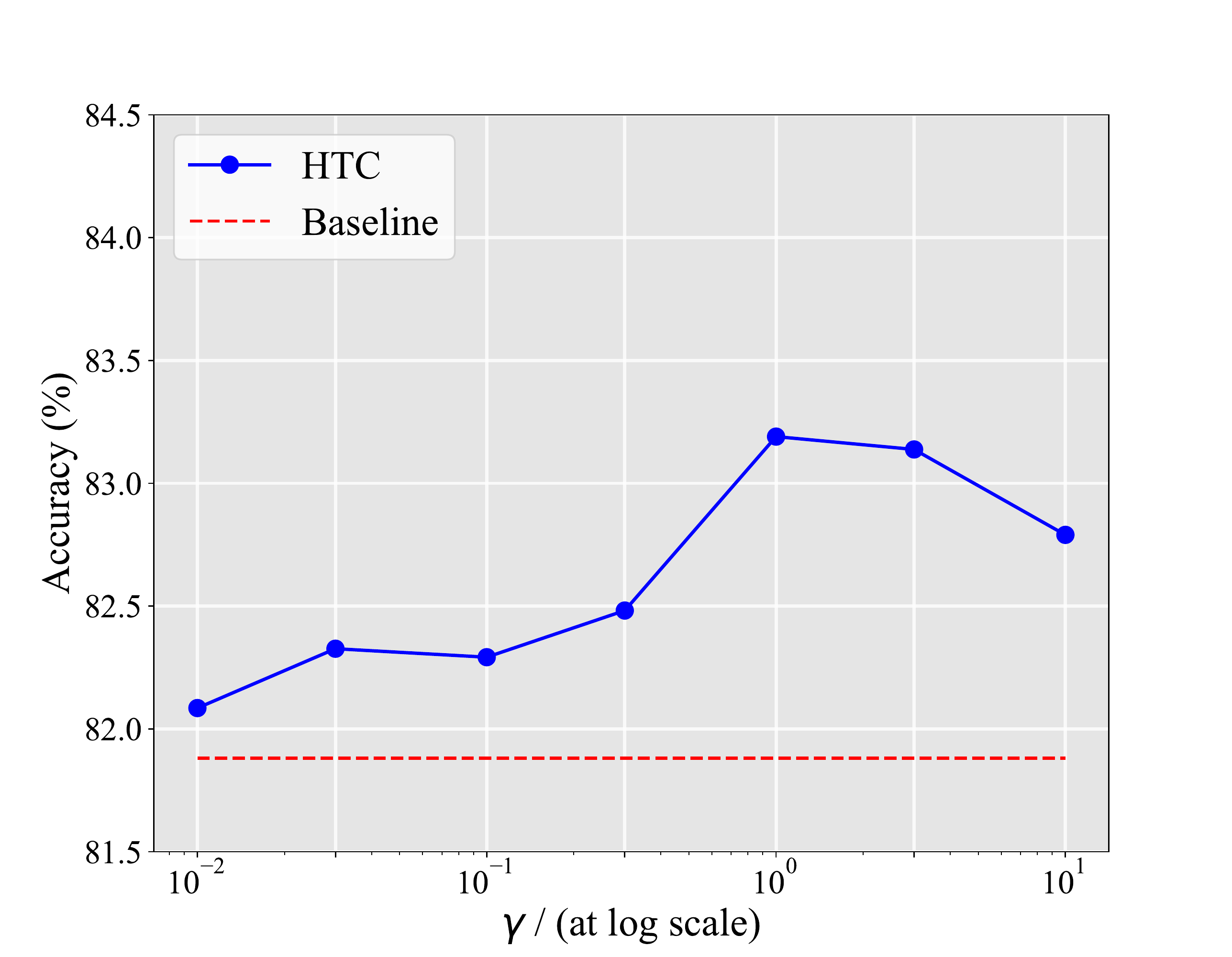}
            \label{fig:gamma}}
    \subfigure[Impact of $\lambda$]{
            \includegraphics[width=0.31\linewidth, trim=5 0 60 50,clip]{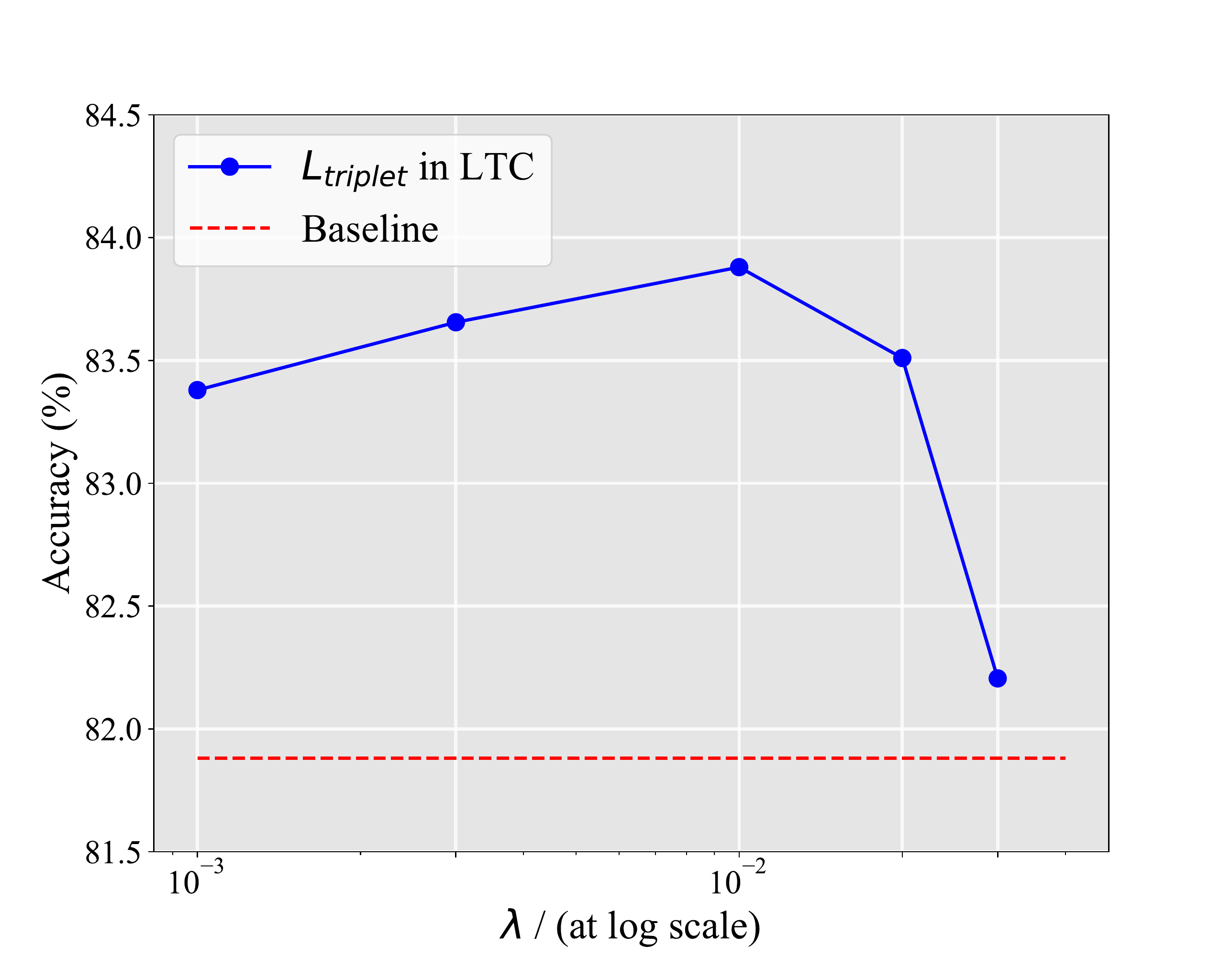}
            \label{fig:lambda}}
    \subfigure[Impact of $\beta$]{
            \includegraphics[width=0.31\linewidth, trim=5 0 60 50,clip]{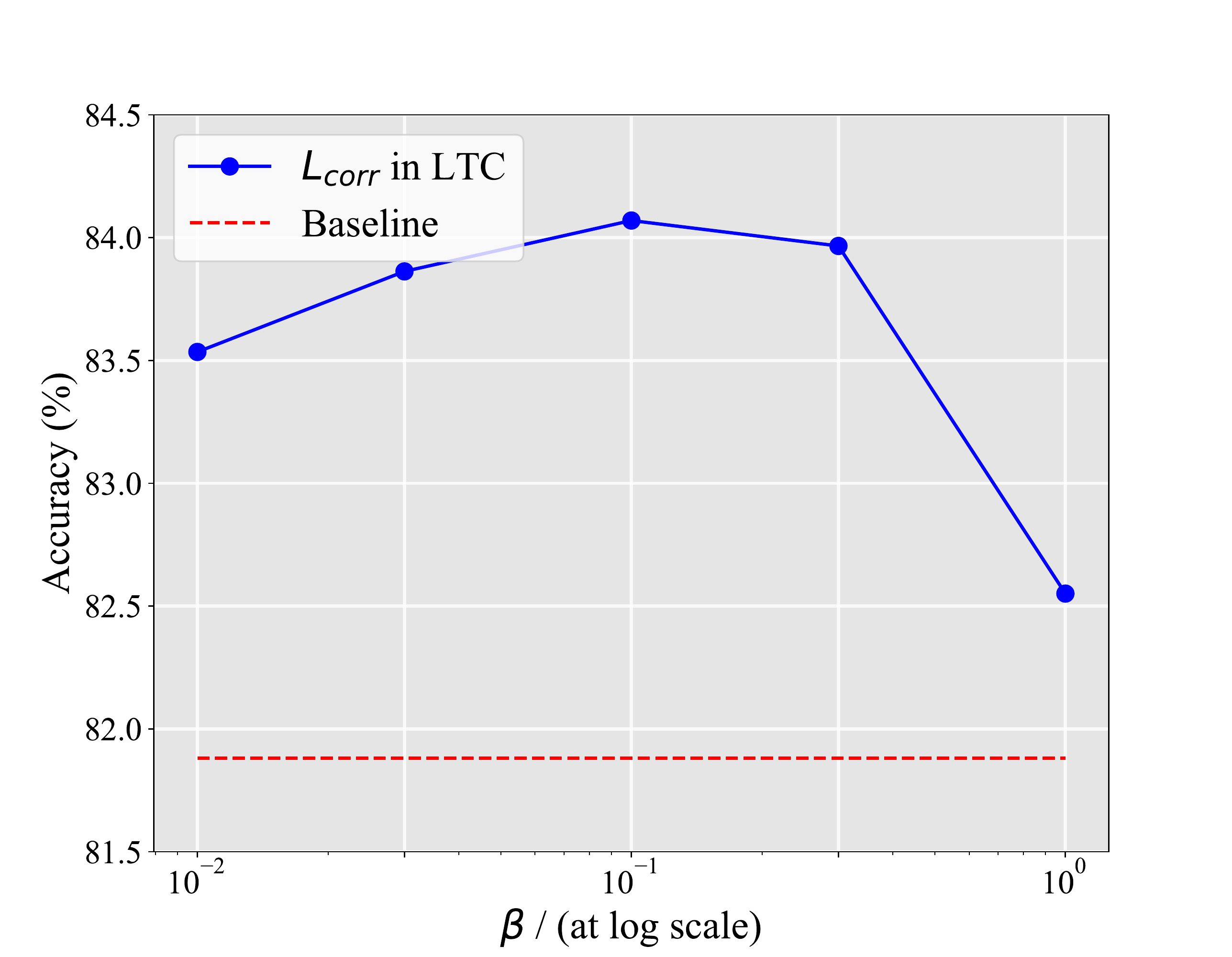}
            \label{fig:beta}}
    \caption{{Effects of hyper-parameters for the HTC constraint with an auxiliary MSE loss, and the margin-based triplet loss $L_\text{triplet}$ and the target correlation consistency loss $L_\text{corr}$ in our proposed LTC method.}}\vspace{-0.2cm}
    \label{fig:parameter}
\end{figure*}

\vspace{0.1cm}\noindent
{ 
\textbf{Effects of the hyper-parameters.} 
To ensure that the proposed components are better trained, we conduct a series of experiments by setting each hyper-parameter to a range of values while fixing the others. 
{Experiments are conducted on the CUB dataset with ResNet18 backbone and default experimental settings for classification.}
We first explore the influence of hyper-parameter $\gamma$ in \textbf{Eq.} (\ref{eq:hadamard_codes}) for the HTC constraint, and vary $\gamma$ from $1e-2$ to $1e+1$. 
As shown in Fig. \ref{fig:gamma}, we can observe that when we set $\gamma=1$, the best result can be gained for the HTC method, so we fix the lambda value with $\gamma=1$ for other experiments both in the HTC and LTC methods. 
Then, we explore the influence of hyper-parameters $\lambda$ and $\beta$ for the margin-based triplet loss in \cref {eq:triplet_loss} and the target correlation consistency loss in \cref {eq:correlation_loss} respectively. 
Similarly, we vary $\lambda$ from $1e-3$ to $1e-1$ and $\beta$ from $1e-2$ to $1$ by controlling the regularization loss in a balanced level, and the results are presented in the Fig. \ref{fig:lambda} and \ref{fig:beta}. 
Absolutely, when we set $\lambda=0.01$ and $\beta=0.1$ in the LTC method, the model can gain the best results respectively. 
Based on those results, we choose the hyper-parameter values leading to the best performance for others.
}

\vspace{0.1cm}\noindent\textbf{Effects of the Length of Target Codes.}
To explore the effects of different lengths of target codes for the HTC and our proposed LTC, we set up two groups of comparative experiments about the length of target codes for the HTC regularization and the LTC regularization, respectively. 
Considering that the length of Hadamard target codes is must larger than the number of categories, the length scale is set from 256 to 4096 in a multiple of 2. For comparison, the length scale for the LTC regularization is the same as that in the HTC. 
Except for the length of the target code, other hyper-parameters are all at the same setting as introduced in the \cref {subsec:exp-detials}. 
We similarly conduct experiments on the CUB dataset with ResNet18 backbone.
The results are presented in \Cref{table:Ablation-Length}, where we can observe that the length of the HTC is more limited by the number of categories and becomes extremely redundant when the length of the HTC is much larger than the number of categories. 
As results showed, the performance would significantly decrease when the length becomes large in the HTC regularization.
On the other hand, the proposed LTC is much more flexible in the length of target codes than the HTC. 
The best results can be achieved when we adopt the LTC regularization with a code length of 512 for the CUB dataset. 
We can also find out that the results for LTC with code lengths from 256 to 4096 can consistently be better than the baseline and the HTC regularization. 
Therefore, the code length is always set as 512 in our other experiments.

\begin{table}[htp]
\centering
\caption{{Effects of the length of target code for the proposed HTC and LTC with ResNet18 Backbone on the CUB dataset. The results are presented in classification accuracy.}} 
\label{table:Ablation-Length}
\resizebox{0.48\columnwidth}{!}{
\begin{tabular}{@{}l|lllll@{}}
\toprule
Length ($L$) & 256   & 512            & 1024  & 2048  & 4096  \\ \midrule
HTC          & 83.19 & 83.19          & 82.43 & 82.27 & 82.39 \\
LTC (Ours)   & 83.66 & \textbf{84.07} & 83.66 & 83.59 & 82.83 \\ \bottomrule
\end{tabular}}\vspace{-0.2cm}
\end{table}

\begin{table}[t]
\centering
\caption{{Effects of the activation function type for the proposed LTC. 
Experiments are conducted on the CUB, the CAR, and the AIR datasets with ResNet18 backbone for classification {(accuracy)}.}} 
\label{table:Ablation-activation}
\resizebox{0.46\columnwidth}{!}{
\begin{tabular}{@{}lccc@{}}
\toprule
Activation   & CUB (\%)       & CAR (\%)       & AIR (\%)       \\ \midrule
tanh ($\xi$=1)   & 83.17          & 91.95          & 89.89          \\
tanh ($\xi$=10)  & 83.88          & 92.50          & 90.58          \\
tanh ($\xi$=100) & 83.83          & \textbf{92.68} & 90.34          \\ \midrule
ste\_sign    & 83.76          & 92.63          & 90.46          \\
sign (ours)  & \textbf{84.07} & \textbf{92.67} & \textbf{90.91} \\ \bottomrule
\end{tabular}}
\end{table}

\vspace{0.1cm}\noindent\textbf{Effects of the activation functions.}
{As previously mentioned in Sec. \ref{subsec:LTC}, due to the vanilla sign function being non-differentiable, we adopt the approximate derivative for the sign function as illustrated in \cref {eq:sign-derivative}. 
For the STE approximate of sign gradient (ste\_sign) proposed in \cite{bengio2013estimating}, the gradient of $sgn(x)$ can be calculated by $\nabla\cdot\frac{\partial \text{sgn}(\bm{w}_k)}{\partial \bm{w}_k} = \nabla$ directly.}
Meanwhile, some existing works \cite{yang2021learning, cao2017hashnet} proposed to adopt the differentiable tanh function for approximating the sign function.
{For comparison, we conduct enough experiments about the tanh function with different scale $\xi$ and the sign function with this constrained approximate derivative for learnable target codes, respectively.}
The experiments are conducted on the CUB, the CAR, and the AIR datasets with ResNet18 backbone for classification. 
{Except for the activation type, all other hyper-parameters are set to the same default values as previously mentioned.}
\Cref{table:Ablation-activation} shows that the best result can always be achieved by using the sign function with this constrained approximate derivative.
As a result, we similarly adopt the sign function with the approximate derivative introduced in \cite{courbariaux2016binarized} for other experiments of LTC. 

\subsection{Visualization}

\begin{figure*}[htbp]
    \centering
    \subfigure[Baseline]{
            \includegraphics[width=0.31\linewidth, trim=30 30 40 50,clip]{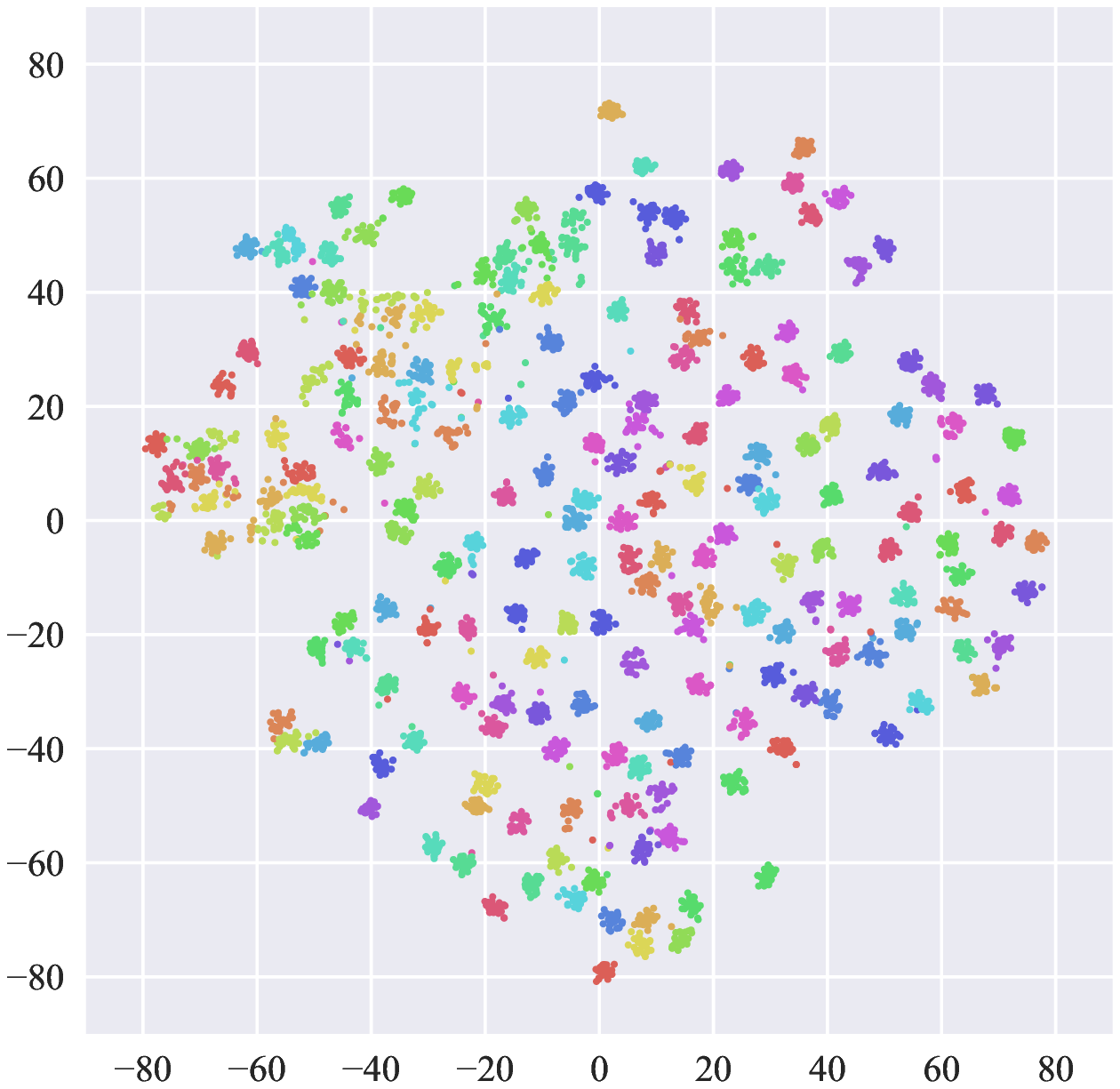}
            \label{fig:subfigure1}}
    \subfigure[the HTC]{
            \includegraphics[width=0.31\linewidth,trim=30 30 40 50,clip]{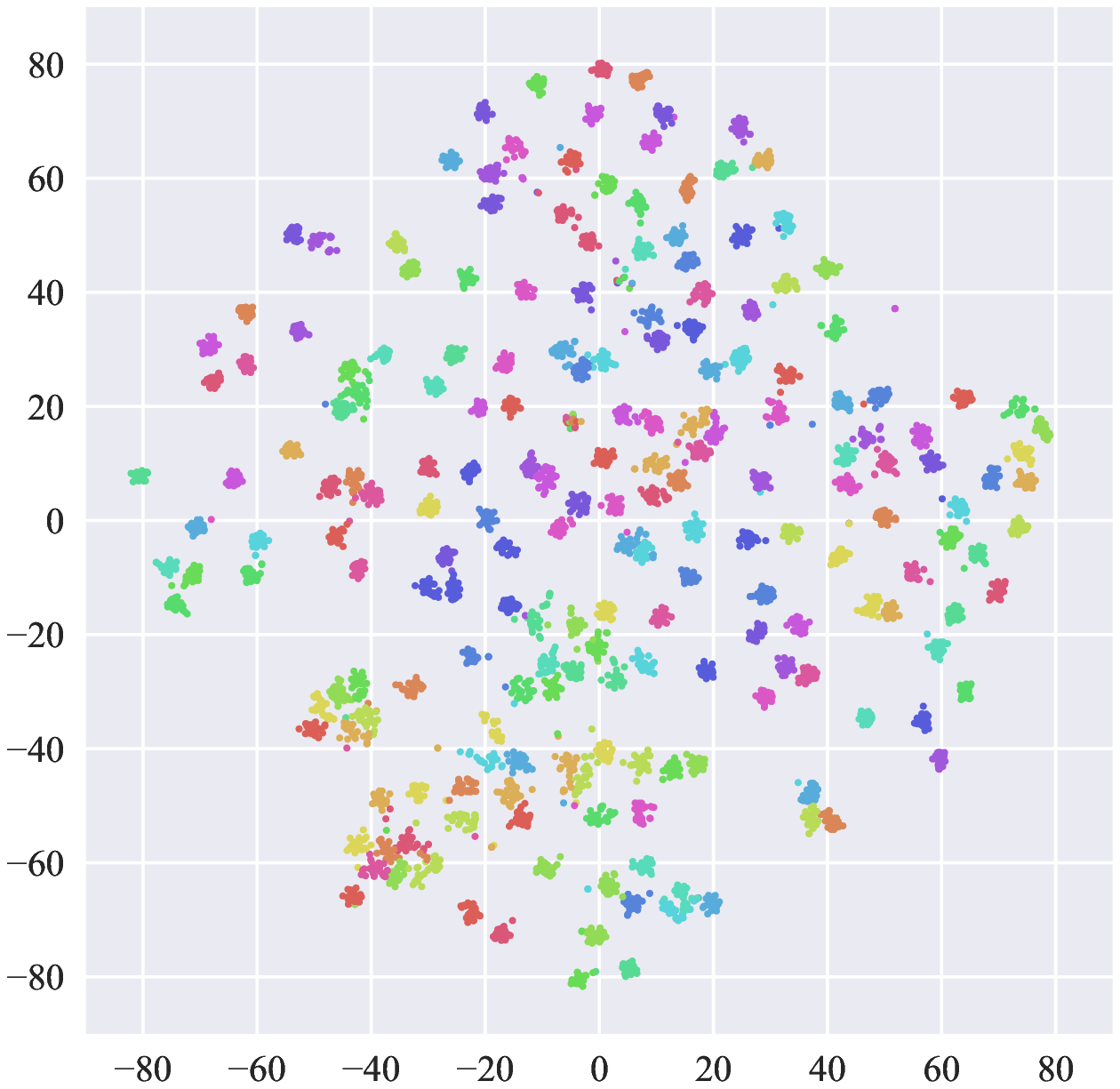}
            \label{fig:subfigure2}}
    \subfigure[the LTC (Ours)]{
            \includegraphics[width=0.31\linewidth, trim=30 30 40 50,clip]{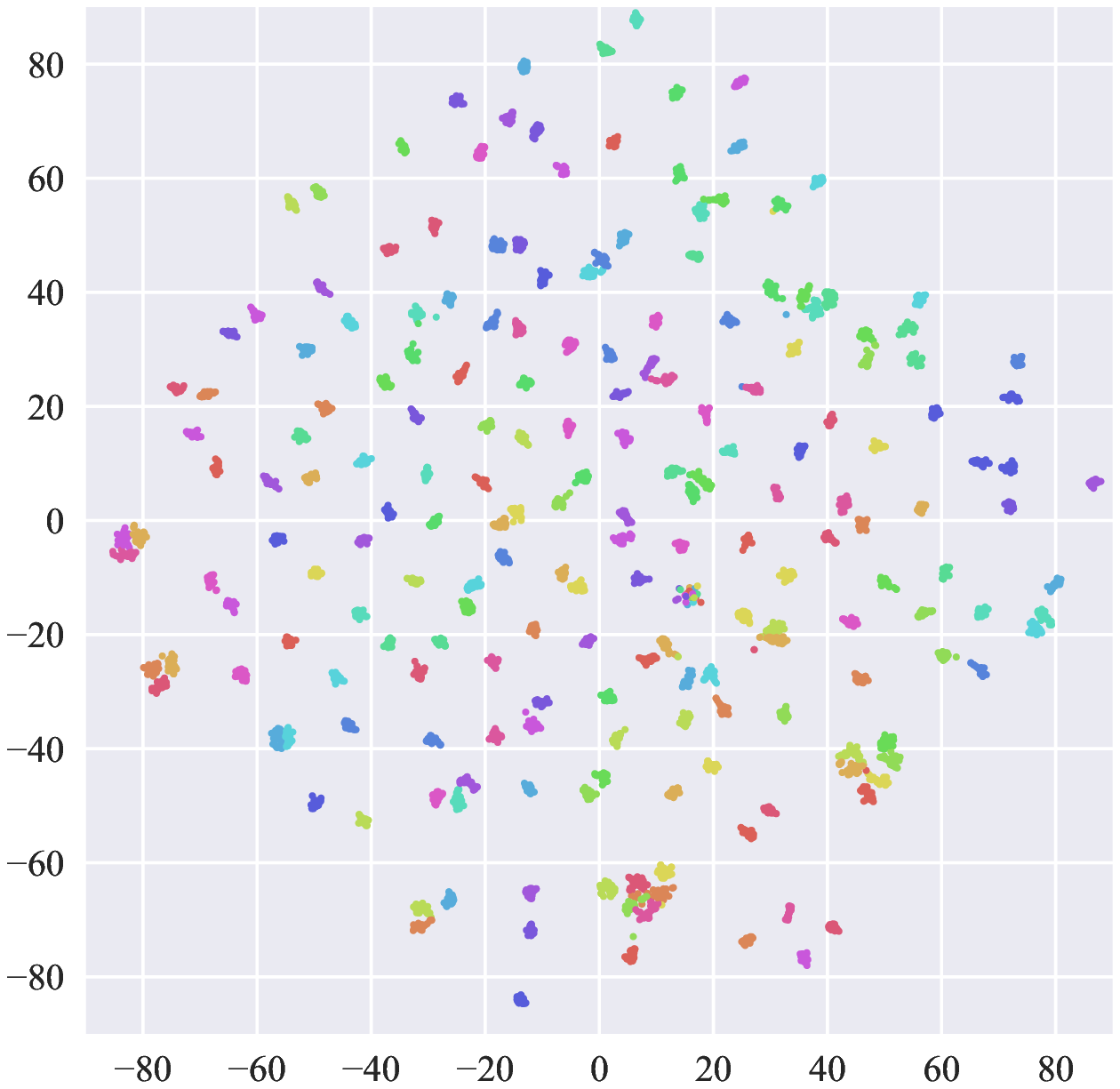}
            \label{fig:subfigure3}}    
    \caption{{Visualization for the baseline, the HTC, and our LTC on the CUB training dataset with the t-SNE \cite{Maaten2008VisualizingDU}. The features for visualization are from the last block of the ResNet-18 backbone of all three models. Note that we use the same color for every 20 classes and visualize the representations of all 200 classes.}}\vspace{-0.2cm}
    \label{fig:visualization_eps}
\end{figure*}

As introduced in \cref{sec:introduction}, the proposed LTC can effectively discover inter-class relations for gaining better representations in supervised learning.
To show the ability of our proposed LTC on the representation learning, we project the features from the same last block of trained models into 2d space by utilizing the t-SNE \cite{Maaten2008VisualizingDU} for three different methods, \ie the baseline, the HTC, and our LTC.
As presented in \cref{fig:visualization_eps}, feature representations of each class with the LTC regularization will be more compact and structured when compared with the Baseline and the HTC method. 
In our analysis, the proposed LTC regularization improves deep representation learning by promoting more structured and compact feature representations for each class by enforcing the geometric properties of target codes. Furthermore, LTC can enhance the discriminative power of learned representations by helping models learn more inter-class relations.

\begin{figure}[t]
  \centering
\includegraphics[width=0.8\linewidth, trim=30 50 5 80,clip]{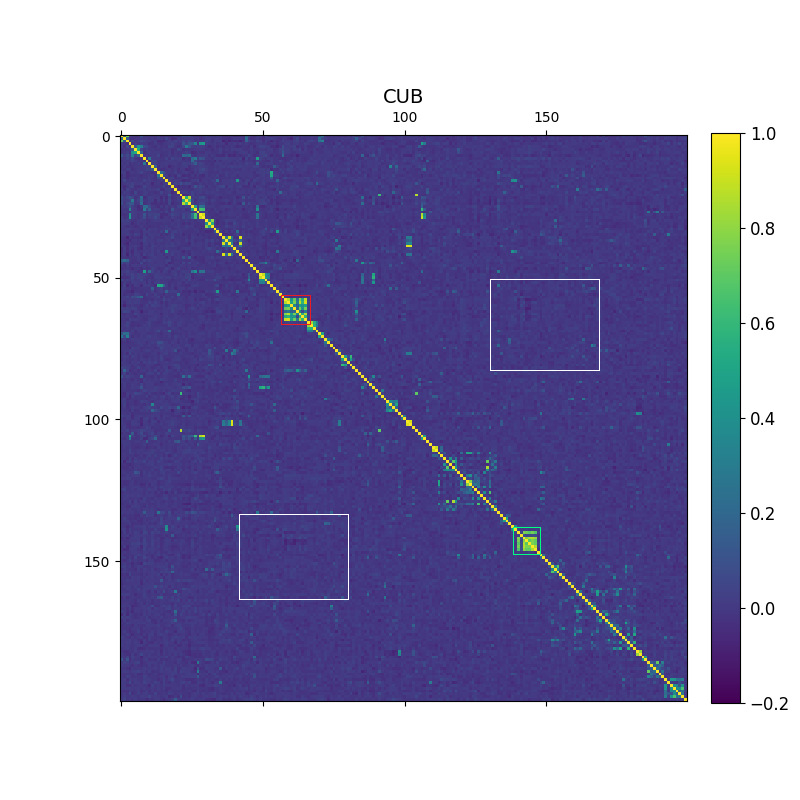}
  \caption{Visualization for the semantic correlations of the learned target codes on the CUB dataset. 
  As illustrated, the semantic correlations in the red bounding box are generated from eight different Gulls with classes id 59-66 in the CUB dataset, while those in the green one are from seven different Terns with classes id 141-147 as showed in \cref{fig:figure1}.
  The brightness of the color represents the magnitude of the correlation, where brighter colors indicate a greater correlation.
  Evidently, the semantic correlations in the red and green bounding boxes are always greater than those in the white ones, which means that our proposed target codes can learn the class-wise correlation well in a weakly supervised way.
  (Best viewed in color)}
  \label{fig:visualization}
\end{figure}

To illustrate the rationale of the LTC, we present the semantic correlations of the learnable target codes as that defined in \cref {subsec:semantic-correlation}. 
Visualization is shown in \cref{fig:visualization}, where the semantic correlation values are normalized by the length of the target code. 
It's observed that categories that are semantically related tend to have stronger correlations with each other compared to those that are less related. This can be seen in \cref{fig:visualization} where the red and green bounding boxes, which represent semantically related categories, have a stronger correlation compared to the white bounding boxes, which represent less related categories.
By leveraging such learnable target codes, the model can focus on learning more class-wise correlated information and avoid irrelevant knowledge, leading to making the representation learning more efficient and robust.
\vspace{-0.2cm}

\section{Conclusion}\label{sec:conclusion}\vspace{-0.2cm}
In this paper, we present an auxiliary target coding regularization method for deep representation learning, employing hand-crafted and learnable target codes respectively. By integrating these target codes into the training process, our method encourages the model to capture richer semantic relations among different classes and enhances learning with imbalanced data by augmenting the representation distance between majority and minority classes. Extensive experiments conducted on widely-used fine-grained classification and imbalanced data learning datasets consistently showcase the superior performance of our method compared to recent approaches, including state-of-the-art methods.
However, a limitation of the LTC is their dependence on labeled data for supervision. In the future, we aim to explore the impact of the LTC on self-supervised and unsupervised learning tasks, which operate without labeled data.
Moreover, although the LTC serves as a regularization term to guide model training, our current implementation does not fully utilize it as an independent supervisory signal. Instead, the LTC primarily serves as an auxiliary constraint rather than directly guiding the model toward specific semantic features or class boundaries. Further exploration is warranted to unlock the complete potential of LTC as a robust and informative supervisory signal.

\section*{Acknowledgment}
This work is supported in part by the Guangdong Pearl River Talent Program (Introduction of Young Talent) under Grant No. 2019QN01X246, the Guangdong Basic and Applied Basic Research Foundation under Grant No. 2023A1515011104 and the Major Key Project of Pengcheng Laboratory under Grant No. PCL2023A08. 
 

\begin{small}
	\bibliographystyle{elsarticle-num}
	\bibliography{egbib}
\end{small}

\end{document}